\documentclass[journal]{IEEEtran}
\IEEEoverridecommandlockouts
\usepackage{cite}
\usepackage{amsmath,amssymb,amsfonts}
\usepackage{algorithmic}
\usepackage{graphicx}
\usepackage{textcomp}
\usepackage{xcolor}
\usepackage{hyperref}
\usepackage{subcaption}
\usepackage{xcolor}
\usepackage{algorithm}
\hypersetup{hidelinks} 
\usepackage{caption}
\def\BibTeX{{\rm B\kern-.05em{\sc i\kern-.025em b}\kern-.08em
    T\kern-.1667em\lower.7ex\hbox{E}\kern-.125emX}}
\begin{document}

\title{Dynamic Scheduling for Vehicle-to-Vehicle Communications Enhanced Federated Learning
}
\author{\IEEEauthorblockN{Jintao Yan, Tan Chen,~\IEEEmembership{Student Member,~IEEE,} Yuxuan Sun,~\IEEEmembership{Member,~IEEE},\\ Zhaojun Nan,~\IEEEmembership{Member,~IEEE,} Sheng Zhou,~\IEEEmembership{Senior Member,~IEEE} and Zhisheng Niu,~\IEEEmembership{Fellow,~IEEE} \vspace{-2.5em}}\\
\thanks{
This work of J. Yan, T. Chen, Z. Nan, S. Zhou and Z. Niu is supported in part by the China Postdoctoral Science Foundation under Grant 2023M742011, and in part by the Project of Tsinghua University-Toyota Joint Research Center for AI Technology of Automated Vehicle under Grant TTAD2025-08. The work of Y. Sun is supported in part by the National Natural Science Foundation of China (No. 62301024, No. U2468201), the Beijing Natural Science Foundation (No. L222044), the Young Elite Scientists Sponsorship Program by CAST, and the Talent Fund of Beijing Jiaotong University under grant 2023XKRC030.

J. Yan, T. Chen, Z. Nan, S. Zhou (Corresponding Author) and Z. Niu are with the Beijing National Research Center for Information Science and Technology, Department of Electronic Engineering, Tsinghua University, Beijing 100084, China. (email: \{yanjt22, chent21\}@mails.tsinghua.edu.cn, nzj660624@mail.tsinghua.edu.cn, \{sheng.zhou, niuzhs\}@tsinghua.edu.cn). 

Y. Sun is with the School of Electronic and Information Engineering, Beijing Jiaotong University, Beijing 100044, China. (e-mail: yxsun@bjtu.edu.cn).}
}

\maketitle

\maketitle


\begin{abstract}
Leveraging the computing and sensing capabilities of vehicles, vehicular federated learning (VFL) has been applied to edge training for connected vehicles. The dynamic and interconnected nature of vehicular networks presents unique opportunities to harness direct vehicle-to-vehicle (V2V) communications, enhancing VFL training efficiency. In this paper, we formulate a stochastic optimization problem to optimize the VFL training performance, considering the energy constraints and mobility of vehicles, and propose a V2V-enhanced dynamic scheduling (VEDS) algorithm to solve it. The model aggregation requirements of VFL and the limited transmission time due to mobility result in a stepwise objective function, which presents challenges in solving the problem. We thus propose a derivative-based drift-plus-penalty method to convert the long-term stochastic optimization problem to an online mixed integer nonlinear programming (MINLP) problem, and provide a theoretical analysis to bound the performance gap between the online solution and the offline optimal solution. Further analysis of the scheduling priority reduces the original problem into a set of convex optimization problems, which are efficiently solved using the interior-point method. Experimental results demonstrate that compared with the state-of-the-art benchmarks, the proposed algorithm enhances the image classification accuracy on the CIFAR-10 dataset by $4.20\%$ and reduces the average displacement errors on the Argoverse trajectory prediction dataset by $9.82\%$.


\end{abstract}
\section{Introduction}

The rapid advancement of vehicular networks has enabled various new applications, including vehicular cooperative perception, trajectory prediction, and route planning. These applications produce vast amounts of data and require timely training of machine learning (ML) models to adapt to changing road conditions \cite{sun2020edge}. In conventional ML frameworks, data is transmitted to a central server for model training, which poses privacy risks and incurs significant delays. As more and more vehicles are equipped with powerful computing capabilities and can collect data via on-board sensors, the ML training process can shift from centralized servers to the vehicles themselves. Therefore, vehicular federated learning (VFL) is a promising framework for timely training and privacy conservation \cite{yan}.

VFL is a distributed ML framework, where an ML model is trained over multiple vehicles. Vehicles with local data and computing capabilities are called \emph{source vehicles} (SOVs). Each SOV trains an ML model based on the local dataset and uploads the model parameters to the roadside unit (RSU). The RSU aggregates the received parameters to obtain a global model and then broadcasts the new models to vehicles to start a new round. Implemented in vehicular networks, VFL takes advantage of the distributed data and processing capabilities while maintaining data privacy \cite{VFLS, VFLS2}.


The distinguished characteristic of VFL is the high mobility of vehicles \cite{VFLS2}, bringing about challenges and opportunities. On the one hand, mobility leads to many challenges. Firstly, the channel conditions of vehicular networks change rapidly due to the high mobility, which complicates the channel estimation and leads to unreliable data transmissions \cite{VFL9}. Secondly, the connections between vehicle-to-infrastructure (V2I) are intermittent. One vehicle may leave the coverage of an RSU before uploading all of the local model parameters \cite{mobfl}, which imposes stringent latency requirements for model aggregation in VFL. The current solution to this problem is to increase the processor frequency and transmission power to reduce the computation and communication latency \cite{VFL7}. However, this may greatly increase the energy consumption of SOVs.

On the other hand, mobility also brings about \emph{communication opportunities} \cite{MEET, chentan}. Recent advancements in vehicle-to-vehicle (V2V) communications via sidelinks enable vehicles to communicate directly with each other, enhancing transmission rates and reliability in vehicular networks \cite{V2Vstd, V2Vstd2}. Many vehicles that are not scheduled for training can also be involved in VFL by \emph{relaying the model uploads}, which are namely \emph{opportunistic vehicles} (OPVs). Utilizing the sidelinks, SOVs can upload their model parameters to the RSUs with the help of OPVs. Mobility increases the likelihood of scheduled vehicles encountering OPVs at closer ranges, under better channel conditions, or with line-of-sight paths. Leveraging these OPVs may increase the success rate of model uploading and therefore enhance the learning performance. 

Currently, many studies have leveraged V2V sidelinks to support various applications in vehicular networks, such as vehicular task offloading \cite{V2Vedge1, V2Vedge2, V2Vedge3}, vehicular edge caching \cite{Vcache1, Vcache2} and cooperative perception \cite{V2Vcp1, V2Vcp2, V2Vcp3}. However, few works utilize V2V sidelinks to improve the performance of VFL. Different from other applications \cite{V2Vedge1, V2Vedge2, V2Vedge3, Vcache1, Vcache2, V2Vcp1, V2Vcp2, V2Vcp3}, VFL operates on a longer time scale with model aggregation requirements. Therefore, a dynamic scheduling algorithm is needed to adapt to the changing environment throughout the VFL training.

In this work, we consider a VFL system that utilizes the V2V communication resources and employs the OPVs to assist SOVs in model uploading, enhancing the VFL performance. The main contributions are summarized as follows:

\begin{itemize}
\item We characterize the convergence bound of the VFL system, and formulate a stochastic optimization problem to minimize the global loss function, considering the energy constraints and the channel uncertainty caused by vehicle mobility. A V2V-enhanced dynamic scheduling (VEDS) algorithm is proposed to solve it.

\item The model aggregation requirements and the limited transmission time in VFL result in a stepwise objective function, which is non-convex and hard to solve. We propose a \emph{derivative-based drift-plus-penalty} method to convert the long-term stochastic optimization problem to an online mixed integer nonlinear programming (MINLP) problem. We provide a theoretical performance guarantee for the proposed transformation by bounding the performance gap between the online and offline solutions. Our analysis further shows the impact of approximation parameters on the performance bound.

\item Through the analysis of the MINLP problem, we identify the \emph{priority} in the OPV scheduling and reduce the original problem to a set of convex optimization problems, which are solved using the interior-point method.

\item Experimental results show that, compared with the state-of-the-art benchmarks, the test accuracy is increased by $3.18\%$ for image classification on the CIFAR-10 dataset, and the average displacement error (ADE) is reduced by $10.21\%$ for trajectory prediction on the Argoverse dataset.

\end{itemize}

The rest of this paper is organized as follows. The related papers are reviewed in Section II. Section III introduces the system model, including the FL, computation, and communication models. The convergence analysis and problem formulation are provided in Section IV, and the VEDS algorithm is proposed in Section V. Experimental results are shown in Section VI, and conclusions are drawn in Section VII.

\begin{figure}[t!]
\centering
\includegraphics[width=0.48\textwidth]{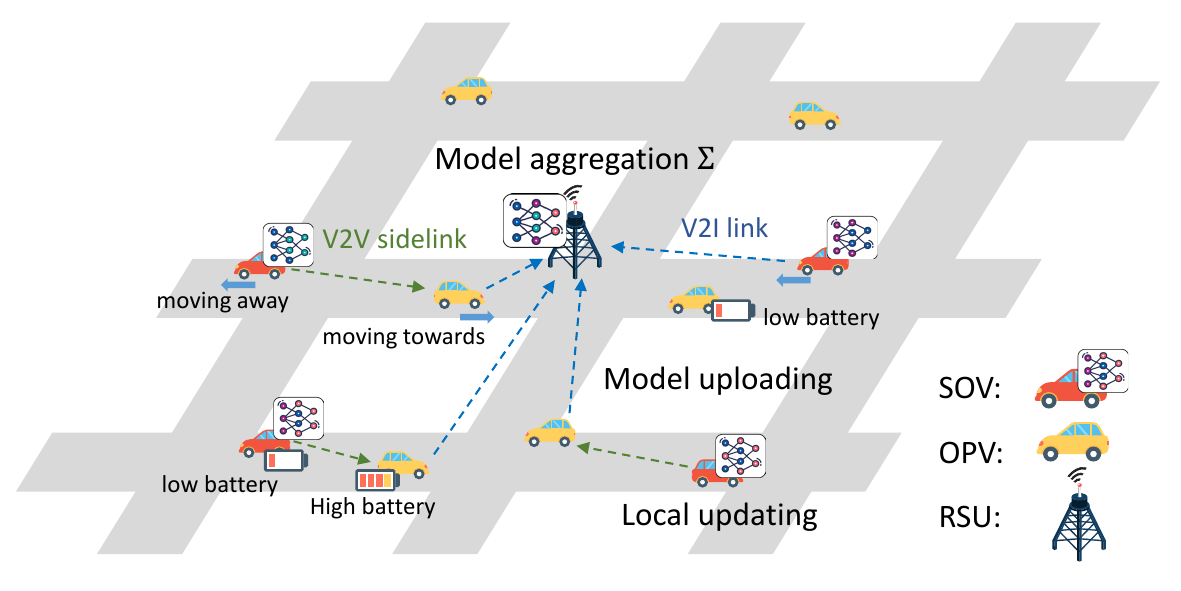}
\caption{The VFL framework.}
\label{system}
\end{figure}




\section{Related Works}
Many studies have explored the application of federated learning (FL) in wireless networks \cite{FLreview}, addressing critical issues such as wireless resource management\cite{DS3, DS4, DS9, DS10, DS6, DS7, DSsun, HHYang}, compression and sparsification\cite{CS4, CS5, CS1, CS2, CS3}, and training algorithm design \cite{FLT1, FLT2, FLT3, Mahdi1, Mahdi2}. However, these studies rarely consider the unique characteristics of vehicular networks, such as high mobility and rapidly changing channel conditions.

More recent studies have begun to investigate FL in vehicular networks. These studies recognize the challenges posed by the high mobility of vehicles and the dynamic nature of vehicular environments \cite{VFL6, VFL9, mobfl, mobfl2,VFL7, VFL8}. In \cite{VFL6}, the impact of vehicle mobility on data quality, such as noise, motion blur, and distortion, is considered, and a resource optimization and vehicle selection scheme is proposed in the context of VFL. The proposed scheme dynamically schedules vehicles with higher image quality, increasing the convergence rate and reducing the time and energy consumption in FL training. In \cite{VFL9}, the short-lived connections between vehicles and RSUs are considered, and a mobility-aware optimization algorithm is proposed. The proposed algorithm enhances the convergence performance of VFL by optimizing the duration of each training round and the number of local iterations. In \cite{mobfl}, a convergence analysis is provided to evaluate the impact of mobile users leaving the coverage area. A mobility-aware solution is proposed to mitigate the impact of mobility by clustering users with similar mobility patterns, which helps stabilize the model updates. Ref. \cite{mobfl2} extends the analysis to the scenarios with heterogeneous data distributions, showing that mobility exacerbates the divergence of data heterogeneity, and therefore degrades the FL convergence. In \cite{VFL7, VFL8}, the impact of rapidly time-varying channels resulting from vehicle mobility is considered. Specifically, a mobility and channel dynamic aware FL (MADCA-FL) scheme is proposed in \cite{VFL7}, which optimizes the success probability of vehicle selection and model parameter updating based on the analysis of vehicle mobility and channel dynamics. In \cite{VFL8}, a more realistic scenario is explored within a 5G new radio framework, and a joint VFL and radio access technology parameter optimization scheme is proposed under the constraints of delay, energy, and cost, aiming to maximize the successful transmission rate of locally trained models. However, most existing studies focus on V2I aggregation, overlooking the potential of harnessing V2V sidelinks to enhance the VFL training efficiency. 

Enhancements in V2V communications through sidelinks, as introduced in the recent updates by the Third Generation Partnership Project (3GPP) \cite{V2Vstd,V2Vstd2}, enable vehicles to communicate with each other directly. This advancement supports a variety of vehicular applications, including vehicular task offloading \cite{V2Vedge1, V2Vedge2, V2Vedge3}, vehicular edge caching \cite{Vcache1, Vcache2} and cooperative perception \cite{V2Vcp1, V2Vcp2, V2Vcp3}. In \cite{V2Vedge1,V2Vedge2}, vehicular task offloading strategies are proposed based on V2V communications, where tasks from one vehicle are offloaded to another to reduce the computational load on the original vehicle and enhance the task execution performance. Further investigations \cite{V2Vedge3} have explored the integration of V2I and V2V communications, utilizing vehicles within the network as relays to improve the efficiency of task offloading processes. In terms of vehicular edge caching, the V2V sidelinks are utilized to enhance the caching hit rate and reduce the content access latency \cite{Vcache1, Vcache2}. In \cite{V2Vcp1, V2Vcp2, V2Vcp3}, the scenario of vehicular cooperative perception is explored, where V2V assistance expands the sensing range and enhances the accuracy of vehicle perception.

In the context of VFL, V2V communication resources have great potential for optimizing training efficiency. By appropriately utilizing these resources, the convergence speed of FL can be significantly improved, and the energy consumption of vehicles can be balanced.



\section{System Model}
\label{II}
\subsection{VFL Model}

We consider a VFL system as shown in Fig. \ref{system}, where an RSU (indexed by $r$ in the following) orchestrates the training of a neural network model $\boldsymbol{w}$ with the assistance of vehicles that enter its coverage area. During the $k^{\text{th}}$ training round, the vehicles that possess local datasets and are willing to participate in the collaborative training of the neural network model are referred to as SOVs, denoted by $\mathcal{S}_k$. The vehicles that do not participate in model training, but have communication capabilities and can help SOVs upload the models are referred to as OPVs, denoted by $\mathcal{U}_k$. The RSU broadcasts control information, including resource allocation and scheduling decisions, while vehicles perform tasks like model uploading or relaying based on these instructions. Vehicles also report status information to the RSU, such as transmitted model parameters, channel conditions, and remaining battery levels.

Each SOV $m \in \mathcal{S}_k$ holds a local dataset with an associated distribution $\mathcal{D}_m$ over the space of samples $\mathcal{X}_m$. For each data sample $\boldsymbol{x} \in \mathcal{X}_m$, a loss function $f(\boldsymbol{w},\boldsymbol{x})$ is used to measure the fitting performance of the model vector $\boldsymbol{w}$. The local loss function of vehicle $m$ is defined as the average loss over the distribution $\mathcal{D}_m$, i.e., $f_m(\boldsymbol{w})\triangleq \underset{\boldsymbol{x}\sim\mathcal{D}_m}{\operatorname*{\mathbb{E}}}[f(\boldsymbol{w},\boldsymbol{x})].$ 

Different from traditional FL, where the set of clients participating in model training is fixed, the set of vehicles participating in VFL training varies in each round due to mobility. We assume that the vehicles are drawn from a given distribution $\mathcal{P}$, and the global loss function is defined as the average local loss function over the distribution $\mathcal{P}$, i.e., 
\begin{equation}F(\boldsymbol{w})\triangleq\underset{m\sim\mathcal{P}}{\operatorname*{\mathbb{E}}}[f_m(\boldsymbol{w})].\label{global}\end{equation} 
The goal is to minimize the global loss function by optimizing the global parameter $\boldsymbol{w}$ through $K$ rounds of training. $\mathcal{K} = \{1,2,..., K\}$ denotes the index of training rounds.

The VFL training process in each round includes three stages: local updates, model uploading and model aggregation. 
\subsubsection{Local Updates} At the start of $k^{\text{th}}$ round, the RSU $r$ broadcasts its model parameters $\boldsymbol{w}_{k-1}$ to the SOVs. After receiving the global model $\boldsymbol{w}_{k-1}$, every SOV $m\in\mathcal{S}_k$ uses stochastic gradient descent (SGD) algorithm to update the local model:
\begin{equation}\boldsymbol{w}_{m,k}=\boldsymbol{w}_{k-1}- \frac{\eta_{k}}{B_{m,k}}\sum_{\boldsymbol{x}\in\mathcal{B}_{m,k}}\nabla f\left(\boldsymbol{w}_{k-1},\boldsymbol{x}\right),\label{GD}\end{equation}
where $\eta_{k}$ is the learning rate, $\mathcal{B}_{m,k}$ is a minibatch randomly sampled from the sample space  $\mathcal{X}_m$. We assume that the batch size of all SOVs is the same, and denote it by $B_{k} = |\mathcal{B}_{m,k}|$.


\subsubsection{Model Uploading} After an SOV completes the local updates, it uploads its model parameters to the RSU for model aggregation. SOVs can upload their model either via a direct V2I link or with the help of the OPVs via a V2V sidelink. The set of SOVs that successfully upload their model to the RSU is denoted by $\hat{\mathcal{S}}_k \in \mathcal{S}_k$. The detailed communication model for model uploading is described in Section \ref{comm}.
\subsubsection{Model Aggregation} At the end of the $k^{\text{th}}$ round, the RSU aggregates the received model parameters:
\begin{equation}\boldsymbol{w}_{k}=\frac{1}{|\hat{\mathcal{S}}_k|}\sum_{m\in \hat{\mathcal{S}}_k} \boldsymbol{w}_{m,k},
\label{agg1}\end{equation}
and then starts a new round.
\subsection{Computation Model}
We adopt a standard computation model \cite{comp1} \cite{comp2} for local updates. The total workload for computing local updates for each vehicle is $N_{\text{flop}} B_{m,k}$, where $N_{\text{flop}}$ is the number of floating point operations (FLOPs) needed for processing each sample. Further, we define $l_{m,k}$ (in cycle/s) as the clock frequency of the vehicular processor in round $k$. Hence, the computation latency for updating the local model is determined as follows:
\begin{equation}
t^{\text{cp}}_{m,k} = \frac{N_{\text{flop}} B_{m,k}}{l_{m,k}},\notag 
\end{equation}
and the computation energy usage is
\begin{equation}
e^{\text{cp}}_{m,k} = \rho l_{m,k}^2N_{\text{flop}} B_{m,k},\notag 
\end{equation}
where $\rho$ is the energy consumption coefficient that depends on the chip architecture of the processor.
\subsection{Communication Model}
\label{comm}
\begin{figure}[t!]
\centering
\includegraphics[width=0.48\textwidth]{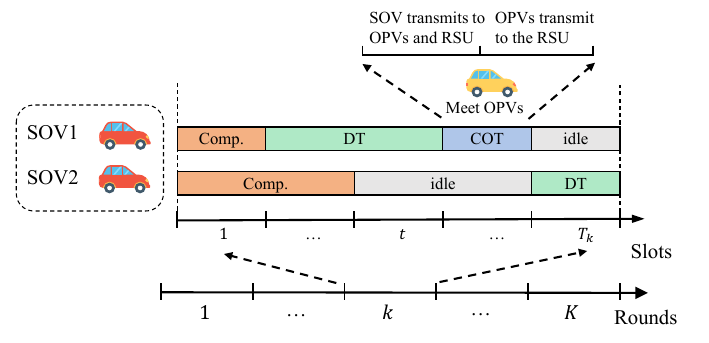}
\caption{The timeline of the proposed VFL system.}
\label{V2V}
\end{figure}

We assume that the vehicular network operates in a discrete time-slotted manner. The slots in round $k$ are denoted by $\mathcal{T}_k = \{1, 2, 3,..., T_k\}$, where $T_k$ is the number of slots in round $k$ and the slot length is denoted by $\kappa$. The round duration $\kappa T_k$ is set to be the average sojourn time of vehicles in the RSU coverage. We assume that, based on historical information, the average sojourn time of vehicles within the RSU coverage area can be estimated, but the specific sojourn time of each vehicle cannot be known in advance. The timeline of the proposed system is shown in Fig. \ref{V2V}.

In every slot, one SOV is scheduled to upload its model parameters to the RSU either via a direct V2I link, called direct transmission (DT), or with the help of the OPVs, called cooperative transmission (COT). We use $\boldsymbol{s}(t)=[s_{1}(t),...,s_{|\mathcal{S}_k|}(t)]$ to denote the SOV scheduling decision. $s_{m}(t) = 1$ if the SOV $m \in \mathcal{S}_k$ is scheduled for model uploading in slot $t$. Otherwise, $s_m(t)=0$. Note that since $t \in \mathcal{T}_k$, the subscript $k$ of $s_{m}(t)$ is omitted for simplicity, and the same applies in the following text. $s_{m}(t)$ has the following constraints:
\begin{equation}
s_{m}(t) \in \{0,1\}, \quad \forall m \in \mathcal{S}_k, \forall t \in \mathcal{T}_k, \label{trans1}\\
\end{equation}
\begin{equation}
\sum_{m \in \mathcal{S}_k} s_{m}(t) \leq 1, \quad \forall t \in \mathcal{T}_k. \label{trans2}
\end{equation}

We use a binary variable $c(t)$ to denote the transmission mode. $c(t) = 0$ if the SOV transmits its model to the RSU via DT. $c(t) = 1$ if the SOV transmits its model to the RSU via COT. $c(t)$ has the binary constraint: 
\begin{equation}
c(t) \in \{0,1\}, \quad \forall t \in \mathcal{T}_k. \label{trans3}\\
\end{equation}

For DT, the scheduled SOV uploads its model parameters to the RSU directly using the whole bandwidth $\beta$. The transmission rate (bit/s) for the SOV $m$ is
\begin{equation}
R_{m}^{\text{DT}}(t) = \beta \log_2 \left(1+\frac{p_m(t)h_{m,r}(t)}{\beta N_0}\right),\notag
\end{equation}
where $h_{m,r}(t)$ is the channel coefficient between vehicle $m$ and the RSU. Due to the high mobility of vehicles, the channel coefficient varies in different slots. If vehicle $m$ leaves the RSU coverage, $h_{m,r}(t)=0$. $p_m(t)$ is the transmission power of vehicle $m$, and $N_0$ is the noise power spectrum density. 

For COT, the scheduled SOV uses the first half of the slot to broadcast its model parameters to the OPVs and the RSU. The OPVs use distributed space-time code (DSTC) \cite{DSTC, coop1, coop2} to relay the model parameters to the RSU in the second half of the slot, as shown in Fig. \ref{V2V}. DSTC is a cooperative communication technique that exploits spatial diversity to enhance transmission reliability by enabling multiple single-antenna OPVs to form a virtual multi-antenna system and improving robustness against channel fading.

We use $\boldsymbol{u}(t) = [u_{1}(t),...,u_{|\mathcal{U}_k|}(t)]$ to denote the OPV scheduling decision in slot $t$, where $u_{n}(t)=1$ if the OPV $n\in\mathcal{U}_k$ is scheduled for COT, and $u_{n}(t)=0$ otherwise. $u_{n}(t)$ has the binary constraint: 
\begin{equation}
u_{n}(t) \in \{0,1\}, \quad \forall n \in \mathcal{U}_k, \forall t \in \mathcal{T}_k. \label{trans4}\\
\end{equation}
The transmission rate of SOV $m$ using COT is \cite{DSTC, coop1, coop2}
\begin{equation}
\begin{aligned}
R_{m}^{\text{COT}}(t) =  \beta \log_2 \bigg(1+ \frac{p_m(t)h_{m,r}(t)}{\beta N_0} \\ + \sum_{n\in\mathcal{U}_k} \frac{u_{n}(t) p_n(t)h_{n,r}(t)}{\beta N_0}  \bigg).\notag
\end{aligned}
\end{equation}
The V2V transmission rate between SOV $m$ and OPV $n$ is 
\begin{equation}
\begin{aligned}
R_{m,n}^{\text{COT-V}}(t) =  \beta \log_2 \bigg(1+ \frac{p_m(t)h_{m,n}(t)}{\beta N_0} \bigg),\notag
\end{aligned}
\end{equation}
where $h_{m,n}(t)$ is the channel coefficient between SOV $m$ and OPV $n$. To ensure that the scheduled OPVs can reliably decode the signal before it begins to transmit, we have the following constraint:
\begin{equation}
\begin{aligned}
& s_{m}(t) c(t) u_{n}(t) R_{m}^{\text{COT}}(t) \leq u_{n}(t) R_{m,n}^{\text{COT-V}}(t), \\ & \forall m \in \mathcal{S}_k, \forall n \in \mathcal{U}_k, \forall t \in \mathcal{T}_k. \label{relay}
\end{aligned}
\end{equation}
We use $\boldsymbol{p}(t) = [p_{1}(t),...,p_{\left(|\mathcal{S}_k| + |\mathcal{U}_k|\right)}(t)]$ to denote the transmission power allocation in slot $t$. There is a power constraint for SOVs:
\begin{equation}0\le p_{m}(t)\le p^{\text{max}}_m,\quad \forall m \in \mathcal{S}_k, \forall t \in \mathcal{T}_k, \label{powerSOV}\end{equation}\
and for OPVs:
\begin{equation}0\le p_{n}(t)\le p^{\text{max}}_n,\quad \forall n \in \mathcal{U}_k, \forall t \in \mathcal{T}_k. \label{power1}\end{equation}
In every slot, the communication energy consumption for each SOV $m \in \mathcal{S}_k$ is 
\begin{equation}  e_m^{\text{cm}}(t) = \kappa p_m(t) \left[s_{m}(t) (1-c(t)) + \frac{1}{2}s_{m}(t)c(t) \right],\label{enSOV}\notag\end{equation}
and for each OPV $n \in \mathcal{U}_k$, it is
\begin{equation}  e_n^{\text{cm}}(t) = \frac{1}{2} \kappa p_n(t) u_{n}(t)c(t).\label{enOPV}\notag\end{equation}
The data transmitted for each SOV $m \in \mathcal{S}_k$ is
\begin{equation}z_m(t) = \kappa \left[s_{m}(t) (1-c(t)) R_{m}^{\text{DT}}(t) + \frac{1}{2}s_{m}(t) c(t) R_{m}^{\text{COT}}(t)\right].\notag\end{equation}
The SOV $m$ has successfully transmitted its model to the RSU if the amount of transmitted model parameters in all slots is greater than or equal to the model size, i.e., $\sum_{t\in \mathcal{T}_k} z_m(t) \geq Q$, where $Q$ denotes the model size. We use an indicator function $\mathbb{I}\left(\sum_{t\in \mathcal{T}_k} z_m(t) \geq Q\right)$ to denote whether the vehicle $m$ has successfully transmitted its model, where $\mathbb{I}(a) = 1$ if condition $a$ is true, and $\mathbb{I}(a) = 0$ otherwise. Using this notation, the aggregation rule (\ref{agg1}) can be rewritten as
\begin{equation}\boldsymbol{w}_{k}=\frac{\sum_{m\in {\mathcal{S}}_k} \mathbb{I}\left(\sum_{t\in \mathcal{T}_k} z_m(t) \geq Q\right) 
 \boldsymbol{w}_{m,k}}{\sum_{m\in \mathcal{S}_k} \mathbb{I}\left(\sum_{t\in \mathcal{T}_k} z_m(t) \geq Q\right) }.
 \label{agg2}\end{equation}

\section{Problem Formulation}
\subsection{Convergence Analysis}
The goal of the VFL is to minimize the global loss function (\ref{global}). However, this objective function is implicit due to the deep and
diverse neural network architectures of ML. Therefore, convergence analysis is performed for an explicit objective function. Following the state-of-the-art literature \cite{DS9, DS10, DSsun, CS4, CS5}, we make the following assumptions:

\noindent \emph{Assumption 1:}
The local loss function $f_m(\boldsymbol{w})$ of each SOV $m$ is $L$-smooth, i.e., 
\begin{equation}
\begin{aligned}f_m(\boldsymbol {w}_{k}&) - f_m(\boldsymbol {w}_{k-1})  \\
&\leq \left < { \nabla f_m(\boldsymbol {w}_{k-1}), \boldsymbol {w}_{k}-\boldsymbol {w}_{k-1} }\right >  + \frac {L}{2} \left \Vert{ \boldsymbol {w}_{k}-\boldsymbol {w}_{k-1} }\right \Vert ^{2}.\notag
\end{aligned}
\end{equation}

\noindent \emph{Assumption 2:}
The stochastic gradient of each SOV $m$ is variance-bounded, i.e.,
\begin{equation}
\underset{\boldsymbol{x}\sim\mathcal{D}_m}{\operatorname*{\mathbb{E}}}\left[\left\| \nabla f(\boldsymbol{w},\boldsymbol{x})-\nabla f_m(\boldsymbol{w})\right\|^2\right]\leq G^2.\notag
\end{equation}
Then, the following Lemma is derived:

\noindent \textbf{Lemma 1.} \emph{Based on the given assumptions and the aggregation rule (\ref{agg2}), the expected loss decreases after one round is upper bounded by}
\begin{subequations}
\begin{align}
&\mathbb{E}[F(\boldsymbol{w}_k)]-\mathbb{E}[F(\boldsymbol{w}_{k-1})]  \leq \eta_k\left(\frac{L\eta_k}2-1\right) \left \Vert{ \nabla F(\boldsymbol {w}_{k-1}) }\right \Vert^2 \notag \\ 
& +\frac{L\eta_k^2}2 \frac{G^2}{B_{k} \sum_{m\in \mathcal{S}_k} \mathbb{I}\left(\sum_{t\in \mathcal{T}_k} z_m(t) \geq Q\right)}.
\tag{12}\label{ubound}
\end{align}
\end{subequations}
\emph{where the expectation is taken over the randomness of SGD.}

\noindent \emph{Proof:} See Appendix \ref{lemma1}. \hfill$\square$ 

We derive the convergence bound for the convex case, assuming that the loss function of each SOV $m$ is $\mu$-strongly convex, i.e.,
\begin{equation}
\begin{aligned}f_m(\boldsymbol {w}_{k}&) - f_m(\boldsymbol {w}_{k-1})  \\
&\geq \left < { \nabla f_m(\boldsymbol {w}_{k-1}), \boldsymbol {w}_{k}-\boldsymbol {w}_{k-1} }\right >  + \frac {\mu}{2} \left \Vert{ \boldsymbol {w}_{k}-\boldsymbol {w}_{k-1} }\right \Vert ^{2}.\notag
\end{aligned}
\end{equation}
Based on Lemma 1, the convergence performance of the proposed VFL after $K$ rounds of training is given by:

\noindent \textbf{Theorem 1.} (Convex) \emph{After $K$ round of training, the difference between $F({\boldsymbol {w}}_{K})$ and the optimal global loss function $F(\boldsymbol {w}^*)$ is upper bounded by}
\begin{subequations}
\begin{align}
&\mathbb {E}[F({\boldsymbol {w}}_{K})]-F(\boldsymbol {w}^*) \notag \\& \leq(\mathbb {E}[F({\boldsymbol {w}}_{0})]-F(\boldsymbol {w}^*))\prod _{k=1}^{K} (1-\mu \eta _{k})\notag\\&+\sum _{k=1}^{K-1} \frac {\eta _{k}}{2}\frac{G^2}{B_{k}\sum_{m\in \mathcal{S}_k} \mathbb{I}\left(\sum_{t\in \mathcal{T}_k} z_m(t) \geq Q\right)} \prod _{j=k+1}^{K}(1-\mu \eta _{k})\notag\\&+\frac {\eta _{K}}{2}\frac{G^2}{B_{K}\sum_{m\in \mathcal{S}_K} \mathbb{I}\left(\sum_{t\in \mathcal{T}_K} z_m(t) \geq Q\right)}. \notag\tag{13}\label{ubound2}
\end{align}
\end{subequations}
\emph{Proof:} See Appendix \ref{Theorem1}. \hfill$\square$

Further, we extend the convergence analysis to cases where the loss function is non-convex, which commonly occurs in ML models such as neural networks. Setting $\eta_k=\eta$ and based on lemma 1, we have the following theorem.

\noindent \textbf{Theorem 2.} (Non-convex) \emph{After $K$ rounds of training, the expected gradient norm of the global loss function is upper bounded by}
\begin{equation}
\begin{aligned}
\frac {1} {K}&\sum_{k=1}^{K} \mathbb{E}\left\|\nabla F(\boldsymbol {w}_{k-1})\right\|^2 \leq \frac{2L\left(F(\boldsymbol{w}_{0})- F(\boldsymbol{w}^{*})\right)}{K^{1/2}}\\
&  + \frac {1} {K^{3/2}}\sum_{k=1}^{K}  \frac{G^2}{B_{k}\sum_{m\in \mathcal{S}_k} \mathbb{I}\left(\sum_{t\in \mathcal{T}_k} z_m(t) \geq Q\right)}.\label{ubound3}
\end{aligned}
\end{equation}
\emph{Proof:} See Appendix \ref{Theorem2}. \hfill$\square$ 

Theorems 1 and 2 show that the convergence bound of VFL improves as the number of successful uploads per round, $\sum_{m \in \mathcal{S}k} \mathbb{I}\left(\sum_{t\in \mathcal{T}_k} z_m(t) \geq Q\right)$, increases. Our algorithm is designed to maximize the number of successful uploads in each round. The number of SOVs in each round $|\mathcal{S}_k|$ in each round serves as an upper limit on successful uploads, which in turn impacts the algorithm performance by constraining the maximum number of successful uploads. 

\subsection{Problem Formulation}
Based on the theorems above, we alternatively minimize the upper bound of $\mathbb {E}[F({\boldsymbol {w}}_{K})]-F(\boldsymbol {w}^*)$ in (\ref{ubound2}) for convex loss functions, and the upper bound of $\frac {1} {K}\sum_{k=1}^{K} \mathbb{E}\left\|\nabla F(\boldsymbol {w}_{k-1})\right\|^2$ in (\ref{ubound3}) for non-convex loss functions. In both cases, this is equivalent to minimizing $\frac{G^2}{B_{k} \sum_{m\in \mathcal{S}_k} \mathbb{I}\left(\sum_{t\in \mathcal{T}_k} z_m(t) \geq Q\right)}$ in each round $k\in \mathcal{K}$. The optimization problem is formulated as

\begin{subequations}
\begin{align}
P0:  &\underset{\boldsymbol{S}_k,\boldsymbol{c}_k,\boldsymbol{U}_k,\boldsymbol{P}_k}{\min} \ \frac{G^2}{B_{k} \sum_{m \in \mathcal{S}_k} \mathbb{I}\left(\sum_{t\in \mathcal{T}_k} z_m(t) \geq Q\right)} \label{obj0} \\
\text{s.t.} \quad &  \sum_{t \in \mathcal{T}_k} e^{\text{cm}}_m(t)+e^{\text{cp}}_{m,k} \leq E^{\text{cons}}_m,\quad \forall m \in \mathcal{S}_k,  \label{energy1}\\
&  \sum_{t \in \mathcal{T}_k} e^{\text{cm}}_n(t) \leq E^{\text{cons}}_n,\quad \forall n \in \mathcal{U}_k,  \label{energy2}\\
& \mathbb{I}\left(t^{\text{cp}}_{m,k} \geq (t-1)\kappa \right) s_{m}(t) = 0, \ \forall m \in \mathcal{S}_k, \forall t \in \mathcal{T}_k,\label{long1}\\
&\text{constraints (\ref{trans1})$-$(\ref{power1}),}\notag
\end{align}
\end{subequations}
where $\boldsymbol{S}_k = [\boldsymbol{s}(1),...,\boldsymbol{s}(T_k)]$ denotes the SOV scheduling, $\boldsymbol{c}_k = [c(1),...,c(T_k)]$ denotes the transmission mode, $\boldsymbol{U}_k = [\boldsymbol{u}(1),...,\boldsymbol{u}(T_k)]$ is the OPV scheduling, $\boldsymbol{P}_k = [\boldsymbol{p}(1),...,\boldsymbol{p}(T_k)]$ is the power allocation throughout round $k$. The constraints (\ref{energy1}) and (\ref{energy2}) indicate that for each vehicle, the total energy consumption cannot exceed the given energy budget. The constraint (\ref{long1}) ensures that the vehicles begin to transmit after they finish local updates. The constraints (\ref{trans1})$-$(\ref{power1}) limit the range of optimization variables.

Since $G$ and $B_{k}$ are constants, minimizing (\ref{obj0}) is equivalent to minimizing $\frac{1}{\sum_{m\in \mathcal{S}_k} \mathbb{I}\left(\sum_{t\in \mathcal{T}_k} z_m(t) \geq Q\right)}$. Also, since $\frac{1}{\sum_{m\in \mathcal{S}_k} \mathbb{I}\left(\sum_{t\in \mathcal{T}_k} z_m(t) \geq Q\right)} > 0$, there is 
\begin{equation}
\begin{aligned}
&\mathop{\arg\min}\limits_{\boldsymbol{S}_k,\boldsymbol{c}_k,\boldsymbol{U}_k,\boldsymbol{P}_k} \frac{1}{\sum_{m\in \mathcal{S}_k} \mathbb{I}\left(\sum_{t\in \mathcal{T}_k} z_m(t) \geq Q\right)} \\& \quad \quad= 
\mathop{\arg\max}\limits_{\boldsymbol{S}_k,\boldsymbol{c}_k,\boldsymbol{U}_k,\boldsymbol{P}_k} \sum_{m\in \mathcal{S}_k} \mathbb{I}\left(\sum_{t\in \mathcal{T}_k} z_m(t) \geq Q\right).
\notag
\end{aligned}
\end{equation}
Therefore, we transform the objective of $P0$ from (\ref{obj0}) to $\max \sum_{m\in \mathcal{S}_k} \mathbb{I}\left(\sum_{t\in \mathcal{T}_k} z_m(t) \geq Q\right)$, and reformulate $P0$ as
\begin{align}
P1:  &\underset{\boldsymbol{S}_k,\boldsymbol{c}_k,\boldsymbol{U}_k,\boldsymbol{P}_k}{\max} \  \sum_{m \in \mathcal{S}_k} \mathbb{I}\left(\sum_{t\in \mathcal{T}_k} z_m(t) \geq Q\right) \label{obj1} \\
\text{s.t.} \quad & \text{constraints (\ref{trans1})$-$(\ref{power1}), (\ref{energy1})$-$(\ref{long1})}.\notag
\end{align}


\section{V2V-Enhanced Dynamic Scheduling Algorithm}
In this section, we propose the VEDS algorithm that solves $P1$ in an online fashion. Firstly, we propose a derivative-based drift-plus-penalty method to convert the long-term stochastic optimization problem into an online MINLP problem. The converted MINLP problem is then decoupled into a DT problem and a COT problem. The DT problem is convex and is directly solved using the Karush-Kuhn-Tucker (KKT) conditions. Analysis of the OPV scheduling priority reduces the COT problem to a set of convex problems, which are solved using the interior-point method.

\subsection{Transformation of the stochastic optimization problem}
$P1$ is a \emph{stochastic optimization problem}. The greatest challenge to solving this problem lies in the uncertainty of channel state information. In vehicular networks, this results from the rapid changes in channels due to the high mobility of vehicles. In reality, future channel information is often difficult to predict, and even if we could acquire future channel information, addressing this problem remains highly complex due to the integer optimization variables and the non-convex objective function. 

One effective way to tackle this kind of problem is the drift-plus-penalty method in Lyapunov optimization \cite{stochastic}\cite{lyp2}. By constructing virtual queues, the long-term stochastic optimization problem is transformed into an online problem and online decision-making algorithms can be designed to solve it. However, the model aggregation requirements and the limited transmission time of VFL result in a stepwise objective function (\ref{obj1}), which cannot be handled by the typical drift-plus-penalty method. Therefore, we propose a derivative-based drift-plus-penalty method to address this challenge. Firstly, we use the shifted sigmoid function to approximate it and transform $P1$ into $P2$.
\begin{subequations}
\begin{align}
P2:  &\underset{\boldsymbol{S}_k,\boldsymbol{c}_k,\boldsymbol{U}_k,\boldsymbol{P}_k}{\max} \ \sum_{m \in \mathcal{S}_k} \sigma \left(\sum_{t\in \mathcal{T}_k} z_m(t)\right)\label{objp2}  \\
\text{s.t.} \quad & \mathbb{I}\left(\zeta_m(t) = Q \right) s_{m}(t) = 0, \ \forall m \in \mathcal{S}_k, \forall t \in \mathcal{T}_k,\label{long2}\\
&\text{constraints (\ref{trans1})$-$(\ref{power1}), (\ref{energy1})$-$(\ref{long1}),}\notag
\end{align}
\end{subequations}
where $\sigma\left(\sum_{t\in \mathcal{T}_k} z_m(t)\right)$ is a shifted sigmoid function, defined as 
$\sigma\left(\sum_{t\in \mathcal{T}_k} z_m(t)\right) \triangleq\left[1+\exp\left(-\alpha\frac{\sum_{t\in \mathcal{T}_k} z_m(t) - Q}{Q}\right)\right]^{-1},$
and $\alpha$ is an approximation parameter. As $\alpha$ increases, the function $\sigma(\cdot)$ converges towards the indicator function $\mathbb{I}(\cdot)$, becoming a more precise approximation. Constraint (\ref{long2}) ensures that a vehicle will not be scheduled after it finishes transmitting its model. 

We define $\boldsymbol{\zeta}(t) = [\zeta_1(t), ..., \zeta_{|\mathcal{S}_k|}(t)]$ as the amount of model parameters that
has been transmitted, where
\begin{equation}
\zeta_m(t)=
\begin{cases}
\min \left(\sum_{\tau = 1}^{t-1} z_m(\tau), Q\right), &\text{for $t > 1$},\\
0, &\text{for $t = 1$},
\end{cases}\label{zeta}
\end{equation}
The derivative of $\sigma(\zeta_m(t))$ with respect to $\zeta_m(t)$ is 
\begin{equation}
\frac{d\sigma (\zeta_m(t))}{d\zeta_m(t)} = \frac{\alpha (1-\sigma(\zeta_m(t))) \cdot \sigma(\zeta_m(t))}{Q}.\notag
\end{equation}
According to (\ref{zeta}), $\zeta_m(t) \in [0,Q]$. As $\zeta_m(t)$ increases from $0$ to $Q$, $\sigma (\zeta_m(t))$ increases from $0$ to $0.5$. Therefore, $\frac{d\sigma (\zeta_m(t))}{d\zeta_m(t)}$ is an increasing function with respect to $\zeta_m(t)$, reaching its minimum when $\zeta_m(t)=0$, and reaching its maximum when $\zeta_m(t)=Q$. We define
\begin{equation}
\psi (\alpha) \triangleq \frac{\partial \sigma (0)}{\partial \zeta_m(t)} \bigg/ \frac{\partial \sigma (Q)}{\partial \zeta_m(t)}.\notag
\end{equation}
$\psi (\alpha)$ is a decreasing function with respect to $\alpha$. Since $\zeta_m(t) \in [0,Q]$, there is
\begin{equation}
\frac{d\sigma (\zeta_m(t))}{d\zeta_m(t)} \geq \psi (\alpha) \frac{\partial \sigma (Q)}{\partial \zeta_m(t)}.\label{psi}
\end{equation}

We convert the long-term stochastic optimization problem into an online optimization problem as follows. For the SOVs, virtual queues $\boldsymbol{q}^{\text{SOV}}(t) = [q_1^{\text{SOV}}(t), ..., q_{|\mathcal{S}_k|}^{\text{SOV}}(t)]$ are created to represent the difference between the cumulative energy consumption up to slot $t$ and the budget, evolving as follows:
\begin{equation}
q_m^{\text{SOV}}(t+1) =\max\left\{q_m^{\text{SOV}}(t)+e_m^{\text{cm}}(t)-\frac{E_m^{\text{cons}}-e_{m,k}^{\text{cp}}}{T_k},0\right\}.\label{queue1}
\end{equation}
Likewise, virtual queues $\boldsymbol{q}^{\text{OPV}}(t) = [q_1^{\text{OPV}}(t), ..., q_{|\mathcal{U}_k|}^{\text{OPV}}(t)]$ are created for the OPVs, evolving as follows: 
\begin{equation}
q_n^{\text{OPV}}(t+1) =\max\left\{q_n^{\text{OPV}}(t)+e_n^{\text{cm}}(t)-\frac{E_n^{\text{cons}}}{T_k},0\right\}.\label{queue2}
\end{equation}
All virtual queues are initialized to 0, i.e., $\boldsymbol{q}^{\text{SOV}}(t)=\boldsymbol{0}$, and $\boldsymbol{q}^{\text{OPV}}(t)=\boldsymbol{0}$. Then, problem $P2$ can be transformed to $P3$:
\begin{subequations}
\begin{align}
&P3:  \underset{\boldsymbol{s}(t),c(t),\boldsymbol{u}(t),\boldsymbol{p}(t)}{\max} V \sum_{m\in \mathcal{S}_k} z_m (t) \frac{d\sigma (\zeta_m(t))}{d\zeta_m(t)}  \notag \\ & \quad \quad  - \sum_{m \in \mathcal{S}_k} q_m^{\text{SOV}}(t) e^{\text{cm}}_m(t)  -\sum_{n \in \mathcal{U}_k} q_n^{\text{OPV}}(t) e^{\text{cm}}_n(t) \label{objp3}\\
\text{s.t.} \ 
& s_{m}(t), c(t), u_{n}(t) \in \{0,1\}, \  \forall m \in \mathcal{S}_k, \forall n \in \mathcal{U}_k, \\
& \sum_{m \in \mathcal{S}_k} s_{m}(t) \leq 1, \label{schedule}\\
& 0\le p_{m}(t)\le p^{\text{max}}_m,\quad \forall m \in \mathcal{S}_k, \label{powercons1}\\
& 0\le p_{n}(t)\le p^{\text{max}}_n,\quad \forall n \in \mathcal{U}_k, \label{powercons2}\\
&s_{m}(t) c(t) u_{n}(t) R_{m}^{\text{COT}}(t) \leq u_{n}(t) R_{m,n}^{\text{COT-V}}(t), \notag \\ &\forall m \in \mathcal{S}_k, \forall n \in \mathcal{U}_k,\\
& \mathbb{I}\left(t^{\text{cp}}_{m,k} \geq (t-1)\kappa \right) s_{m}(t) = 0, \ \forall m \in \mathcal{S}_k,\\
& \mathbb{I}\left(\zeta_m(t) = Q \right) s_{m}(t) = 0, \ \forall m \in \mathcal{S}_k.
\end{align}
\end{subequations}
We derive the following theorem to guarantee the performance of the proposed transformation. Superscript $^\dagger$ is used to denote the solution to $P3$, and $^*$ is used to denote the optimal offline solution to $P2$.

\noindent \textbf{Theorem 3.} \emph{Suppose all queues are initialized to 0, the difference between the optimal value of solving $P2$ and the counterpart of solving $P3$ is bounded by:}
\begin{equation}
\sum_{m\in \mathcal{S}_k} \sigma\left(\sum_{t\in \mathcal{T}_k} z_m^*(t)\right) - \sum_{m\in \mathcal{S}_k} \sigma\left(\sum_{t\in \mathcal{T}_k} z_m^\dagger(t)\right) \leq  \frac{T_k^2 \Phi}{V\psi (\alpha)}.\label{bound}
\end{equation}
\emph{The energy consumption of the SOV $m \in \mathcal{S}_k$ is bounded by}
\begin{equation}
\begin{aligned}
& \sum_{t \in \mathcal{T}_k} e^{\text{cm}}_m(t)+e^{\text{cp}}_{m,k}  \\ &\leq E^{\text{cons}}_m + \sqrt{2T_k^2 \Phi - 2V  \sum_{t\in \mathcal{T}_k} \sum_{m\in \mathcal{S}_k} z_m^* (t) \frac{d\sigma (\zeta_m(t))}{d\zeta_m(t)}},
\end{aligned}
\end{equation}
\emph{and that of the OPV $n \in \mathcal{U}_k$ is bounded by}
\begin{equation}
\begin{aligned}
& \sum_{t \in \mathcal{T}_k} e^{\text{cm}}_n(t) \\ &\leq E^{\text{cons}}_n + \sqrt{2T_k^2 \Phi - 2V  \sum_{t\in \mathcal{T}_k} \sum_{m\in \mathcal{S}_k} z_m^* (t) \frac{d\sigma (\zeta_m(t))}{d\zeta_m(t)}},
\end{aligned}
\end{equation}
\emph{where $\delta_m^{\text{SOV}}(t) \triangleq e^{\text{cm}}_m(t)-\frac{E^{\text{cons}}_m - e^{\text{cp}}_{m,k}}{T_k}$, $\delta_n^{\text{OPV}}(t) \triangleq e^{\text{cm}}_n(t)-\frac{E^{\text{cons}}_n}{T_k}$, $\phi_m^{\text{SOV}} \triangleq \max_t\{|\delta_m(t)|\}$, $\phi_n^{\text{OPV}} \triangleq \max_t\{|\delta_n(t)|\}$, and $\Phi \triangleq \sum_{m\in \mathcal{S}_k}(\phi_m^{\text{SOV}})^2+\sum_{n\in \mathcal{U}_k}(\phi_n^{\text{OPV}})^2$.} 

\noindent \emph{Proof:} See Appendix \ref{Theorem3}. \hfill$\square$

Theorem 3 shows that, instead of solving the long-term stochastic optimization problem $P2$, we alternatively solve the online problem $P3$. The performance is bounded with respect to the optimal offline solution to $P2$, and the energy consumption for each vehicle is also bounded. The trade-off between the objective function (\ref{objp2}) and the energy consumption is balanced by the weight parameter $V$. The worst-case performance can be improved by increasing the parameter $\psi(\alpha)$, equivalent to reducing the approximation parameter $\alpha$. However, choosing overly small $\alpha$ values compromises the precision of approximating the indicator function $\mathbb{I}(\cdot)$ with the sigmoid function $\sigma(\cdot)$. Therefore, in practice, it is crucial to carefully choose the values of $V$ and $\alpha$ to ensure optimal approximation performance under the energy constraints.

We remark here that, the scheduling and resource allocation decisions are mainly based on the amount of model transmitted, channel state, and virtual queues. As the number of vehicles increases or the VFL model size grows, limited communication and energy resources may result in incomplete model transmissions. The derivative-based drift-plus-penalty method prioritizes scheduling vehicles with better channel conditions and those that have already uploaded a large portion of their models. By focusing resources on these vehicles, the algorithm ensures that they can complete their uploads successfully, thereby maintaining overall performance under constrained resources.

$P3$ is an MINLP with binary variables $\boldsymbol{s}(t),c(t),\boldsymbol{u}(t)$ and continuous variables $\boldsymbol{p}(t)$, which exhibits high computational complexity for direct solution. However, due to the existence of constraint (\ref{schedule}), enumerating $\boldsymbol{s}(t)$ and $c(t)$ only introduces a linear increase in computational complexity. Therefore, we fix the SOV scheduling decision $\boldsymbol{s}(t)$ and transmission mode $c(t)$, and focus on solving $\boldsymbol{u}(t)$ and $\boldsymbol{p}(t)$. Specifically, when the SOV scheduling $\boldsymbol{s}(t)$ and transmission mode $c(t)$ are decided, $P3$ is reduced to the following sub-problems.
\begin{algorithm}[!t]	
    \caption{The procedure of solving $P3$} 
    \label{Algo1}
	\begin{algorithmic}
        \STATE{\textbf{Input:} $\mathcal{H}(t)$, $\boldsymbol{q}^{\text{SOV}}(t)$, $\boldsymbol{q}^{\text{OPV}}(t)$ and $\boldsymbol{\zeta}(t)$;}
        \STATE{\textbf{Output:} The solution $\boldsymbol{s}^*(t), c^*(t), \boldsymbol{u}^*(t),\boldsymbol{p}^*(t)$ to $P3$;}
        \STATE{Initialize $\boldsymbol{s}^*(t)=\boldsymbol{0}$, $c^*(t) =0$, $\boldsymbol{u}^*(t)=\boldsymbol{0}$ and $\boldsymbol{p}^*(t)=\boldsymbol{0}$.}
	\FOR{$m$ \textbf{in} $\mathcal{S}_k$}
        \STATE{Set $\boldsymbol{s}(t)=\boldsymbol{0}$;}
        \STATE{Set $s_m(t)=1$;}
        \IF{$t^{\text{cp}}_{m,k} \leq (t-1) \kappa$         \textbf{and} $\zeta_m(t) \neq Q $}
        \STATE{Set $c(t)=0$, $\boldsymbol{u}(t)=\boldsymbol{0}$ and $\boldsymbol{p}(t)=\boldsymbol{0}$;}
        \STATE{Solve $P3.1$ to obtain $\boldsymbol{p}(t)$;}
        \STATE{Take $\boldsymbol{s}(t)$, $c(t)$, $\boldsymbol{u}(t)$ and $\boldsymbol{p}(t)$ to (\ref{objp3}) to get $y(t)$;}
        \IF{$y(t)\geq y^*(t)$}
        \STATE{Set $\boldsymbol{s}^*(t)=\boldsymbol{s}(t)$, $c^*(t) =c(t)$, $\boldsymbol{u}^*(t)=\boldsymbol{u}(t)$, and $\boldsymbol{p}^*(t)=\boldsymbol{p}(t)$;}
        \ENDIF
        \STATE{Set $c(t)=1$;}
        \STATE{Sort the elements of $\mathcal{U}_k$ in descending order based on the values of $h_{m,n}(t)$;}
        \FOR{$i = 1$ \textbf{to} $|\mathcal{U}_k|$}
        \STATE{For the top $i$ elements in $\mathcal{U}_k$, set the corresponding $u_{n}(t)$to $1$. For all other elements, set $u_{n}(t)$ to $0$;}
        \STATE{Solve $P4$ to obtain $\boldsymbol{p}(t)$;}
        \STATE{Taking $\boldsymbol{s}(t)$, $c(t)$, $\boldsymbol{u}(t)$ and $\boldsymbol{p}(t)$ back to (\ref{objp3}) to obtain $y(t)$;}
        \IF{$y(t)\geq y^*(t)$}
        \STATE{Set $\boldsymbol{s}^*(t)=\boldsymbol{s}(t)$, $c^*(t) =c(t)$, $\boldsymbol{u}^*(t)=\boldsymbol{u}(t)$, and $\boldsymbol{p}^*(t)=\boldsymbol{p}(t)$.}
        \ENDIF
        \ENDFOR
        \ENDIF
	    \ENDFOR
	\end{algorithmic}
\end{algorithm}
\subsection{Direct Transmission Problem}
When SOV $m$ is scheduled for transmission and DT mode is selected ($c(t) = 0$), $P3$ is reduced to
\begin{subequations}
\begin{align}
P3.1:  &\underset{p_m(t)}{\max} \ V \frac{d\sigma (\zeta_m(t))}{d\zeta_m(t)} \kappa R_{m}^{\text{DT}}(t) - \kappa q_m^{\text{SOV}}(t)  p_m(t)\\
\text{s.t.} \quad & 0\le p_{m}(t)\le p^{\text{max}}_m.\label{powercons3}
\end{align}
\end{subequations}
$P3.1$ is a convex problem. The optimal solution is derived using the KKT conditions, in Proposition 1:

\noindent \textbf{Proposition 1.} \emph{Given the SOV scheduling decision, the optimal power allocation strategy for DT is given by} 
\begin{equation}
p_m^*(t) = \left[\frac{V \frac{d\sigma (\zeta_m(t))}{d\zeta_m(t)} \beta}{q_{m}^{\text{SOV}}(t)} - \frac{\beta N_0}{h_{m,r}(t)}  \right]^{p^{\text{max}}_m}_0,
\end{equation}
\emph{where} $[a]^{p^{\text{max}}_m}_0$ \emph{is defined as} $\min(\max(a,0), p^{\text{max}}_m)$.

\noindent \emph{Proof:} See Appendix \ref{Proposition1}. \hfill$\square$

\subsection{Cooperative Transmission Problem}
When SOV $m$ is scheduled for transmission and COT mode is selected ($c(t) = 1$), $P3$ is reduced to
\begin{subequations}
\begin{align}
P3.2:  &\underset{\boldsymbol{u}(t),\boldsymbol{p}(t)}{\max} V \frac{d\sigma (\zeta_m(t))}{d\zeta_m(t)}  \frac{1}{2} \kappa R_{m}^{\text{COT}}(t) - \frac{1}{2} \kappa q_m^{\text{SOV}}(t)  p_m(t) \notag  \\& - \sum_{n \in \mathcal{U}_k} \frac{1}{2} \kappa u_{n}(t) q_{n}^{\text{OPV}}(t)  p_n(t) \\
\text{s.t.} \quad & u_{n}(t) \in \{0,1\}, \quad \forall n \in \mathcal{U}_k,\\
& u_{n}(t) R_{m}^{\text{COT}}(t) \leq u_{n}(t) R_{m,n}^{\text{COT-V}}(t), \quad \forall n \in \mathcal{U}_k, \label{relay2} \\
& \text{constraint (\ref{powercons1}), (\ref{powercons2})}. \notag
\end{align}
\end{subequations}
$P3.2$ is still an MINLP problem, and directly enumerating the binary variable $\boldsymbol{u}(t)$ introduces exponential complexity. We further analyze the OPV scheduling priority and prove the following proposition.

\noindent \textbf{Proposition 2.} \emph{Suppose $P3.2$ is solvable, then there must exist an optimal set of $\boldsymbol{u}(t)$ that adheres to a specific structure: the $u_{n}(t)$ variables are arranged according to the descending order of $h_{m,n}(t)$ values, and the optimal solution involves selecting the top $i$ $u_{n}(t)$ based on this ordering.} 

\noindent \emph{Proof:} This proposition is proved by contradiction. Assume that all optimal solutions $\{\boldsymbol{u}^*(t), \boldsymbol{p}^*(t)  \}$ do not adhere to the proposed structure, i.e., they do not select the top $i$ $u_n(t)$ based on the highest $h_{m,n}(t)$ values. 

Consider one of the optimal solutions $\{\boldsymbol{u}'(t), \boldsymbol{p}'(t)  \}$, that includes some $u_n(t)$ with lower $h_{m,n}(t)$ values set to 1, while at least one $u_n(t)$ with a higher $h_{m,n}(t)$ value (within the top $i$) is set to 0. 
Consider another OPV scheduling strategy $\boldsymbol{u}^\dagger(t)$, where all $u_n^\dagger(t)$ within the top $i$ highest $h_{m,n}(t)$ values are set to 1. For all $n \in \mathcal{U}_k$, we consider the power allocation strategy:
\begin{equation}
p_n^\dagger(t)=
\begin{cases}
p'_n(t), \quad &\text{if $u'_{n}(t)=1$},\\
0, \quad &\text{otherwise}, \notag
\end{cases}
\end{equation}
and set $p_m^\dagger(t) = p'_m(t)$. The solution set $\{\boldsymbol{u}^\dagger(t), \boldsymbol{p}^\dagger(t)  \}$ is a feasible solution, since all constraints of $P4$ are satisfied, and the objective function is 
\begin{equation}
\begin{aligned}
&V \frac{d\sigma (\zeta_m(t))}{d\zeta_m(t)} \frac{1}{2} \kappa \beta \log_2 \bigg(1+ \frac{p^\dagger_m(t)h_{m,r}(t)}{\beta N_0}  \\ & + \sum_{n\in\mathcal{U}_k} \frac{u^\dagger_{n}(t) p^\dagger_n(t)h_{n,r}(t)}{\beta N_0}  \bigg) - \frac{1}{2} \kappa q_m^{\text{SOV}}(t)  p^\dagger_m(t) \\&  - \sum_{n \in \mathcal{U}_k} \frac{1}{2} \kappa u^\dagger_{n}(t) q_{n}^{\text{OPV}}(t)  p^\dagger_n(t)\\
& = V \frac{d\sigma (\zeta_m(t))}{d\zeta_m(t)} \frac{1}{2} \kappa \beta \log_2 \bigg(1+ \frac{p'_m(t)h_{m,r}(t)}{\beta N_0}  \\ & + \sum_{n\in\mathcal{U}_k} \frac{u'_{n}(t) p'_n(t)h_{n,r}(t)}{\beta N_0}  \bigg) - \frac{1}{2} \kappa q_m^{\text{SOV}}(t)  p'_m(t) \\&  - \sum_{n \in \mathcal{U}_k} \frac{1}{2} \kappa u'_{n}(t) q_{n}^{\text{OPV}}(t)  p'_n(t). \notag
\end{aligned}
\end{equation}
Since the objective function of the solution set $\{\boldsymbol{u}^\dagger(t), \boldsymbol{p}^\dagger(t)  \}$ is equal to that of the optimal solution set $\{\boldsymbol{u}'(t), \boldsymbol{p}'(t)  \}$, $\{\boldsymbol{u}^\dagger(t), \boldsymbol{p}^\dagger(t)  \}$ is also an optimal solution. This contradicts the assumption that none of the optimal solutions $\{\boldsymbol{u}^*(t), \boldsymbol{p}^*(t)  \}$ adhere to the proposed structure. Proposition 2 is proved.\hfill$\square$

Based on Proposition 2, we can sort the elements of $\mathcal{U}_k$ in descending order based on the values of $h_{m,n}(t)$, and schedule the first $i$ OPVs for COT, i.e., set $u_{n}(t) = 1$ for them, and set $u_{n}(t)=0$ for all other vehicles. 

When $\boldsymbol{u}(t)$ is given, the constraint (\ref{relay2}) becomes
\begin{equation}
\begin{aligned}
& 1+ \frac{p_m(t)h_{m,r}(t)}{\beta N_0} + \sum_{n\in\mathcal{R}(t)} \frac{p_n(t)h_{n,r}(t)}{\beta N_0} \\
& \leq 1+ \frac{p_m(t)h_{m,n}(t)}{\beta N_0}, \quad \forall n \in \mathcal{R}(t), \label{relay3}
\end{aligned}
\end{equation}
where $\mathcal{R}(t) = \{n \mid u_{n}(t) = 1, n\in \mathcal{U}_k\}$. $P3.2$ is reduced to
\begin{align}
P4:  &\underset{\boldsymbol{p}(t)}{\max} \ V \frac{d\sigma (\zeta_m(t))}{d\zeta_m(t)} \frac{1}{2} \kappa \beta \log_2 \bigg(1+ \frac{p_m(t)h_{m,r}(t)}{\beta N_0} \notag \\ & + \sum_{n\in\mathcal{R}(t)} \frac{p_n(t)h_{n,r}(t)}{\beta N_0}  \bigg) -  \frac{1}{2} \kappa q_m^{\text{SOV}}(t) p_m(t) \notag \\&-  \sum_{n \in \mathcal{R}(t)}  \frac{1}{2} \kappa q_{n}^{\text{OPV}}(t)  p_n(t) \label{objp4} \\
\text{s.t.} \quad & \text{constraints (\ref{powercons2})}, \text{(\ref{powercons3})}, \text{(\ref{relay3})}. \notag
\end{align}
$P4$ is a convex optimization problem since the objective (\ref{objp4}) is to maximize a concave function and all constraints (\ref{powercons2}), (\ref{powercons3}) and (\ref{relay3}) are convex, which can be solved by optimization
tools, such as CVX \cite{cvx}, based on the interior-point method. All transformations of $P3$ are equivalent transformations, and the procedure of solving $P3$ is summarized in Algorithm \ref{Algo1}, where $\mathcal{H}(t) = \{h_{m,n}(t)\ | m\in\mathcal{S}_k, n\in\mathcal{U}_k \cup r\}$, and $y(t)$ denotes the value of (\ref{objp3}), i.e., the objective function of $P3$.
\begin{algorithm}[!t]	
    \caption{The procedure of the VEDS algorithm} 
    \label{Algo2}
	\begin{algorithmic}
        \STATE{\textbf{Initialization} Set $\boldsymbol{q}^{\text{SOV}}(1)$, $\boldsymbol{q}^{\text{OPV}}(1)$ and $\boldsymbol{\zeta}(1)$ to $\boldsymbol{0}$;}
	\FOR{$t$ \textbf{in} $\mathcal{T}_k$}
         \STATE{Update the amount of transmitted model parameters $\boldsymbol{\zeta}(t)$ according to (\ref{zeta});}
        \STATE{Observe the current channel state $\boldsymbol{h}(t)$;}
        \STATE{Solve $P3$ to get $\boldsymbol{s}^*(t), c^*(t), \boldsymbol{u}^*(t),\boldsymbol{p}^*(t)$ based on Algorithm \ref{Algo1}, and allocate communication resources;}
        \STATE{Update the virtual queue $\boldsymbol{q}^{\text{SOV}}(t)$ and $\boldsymbol{q}^{\text{OPV}}(t)$ according to (\ref{queue1}) and  (\ref{queue2});}
	    \ENDFOR
	\end{algorithmic}
\end{algorithm}
\subsection{The Complete Algorithm and Discussions}
The whole procedure of the proposed VEDS algorithm is summarized in Algorithm \ref{Algo2}. At the start of each round, the RSU broadcasts the global model to the SOVs, and the SOVs perform local updates based on their local dataset. In each slot, the RSU solves $P3$ based on the current channel state $\boldsymbol{h}(t)$ and the amount of transmitted model parameters $\boldsymbol{\zeta}(t)$. Based on the solution to $P3$, the resources are allocated, and the virtual queues are updated. This process is iterated until the end of the round. 

To further elaborate on the advantages of VEDS in enhancing VFL performance, we discuss its impact from three key aspects. Firstly, dynamic scheduling enables real-time adaptation to vehicular mobility by scheduling OPVs with favorable channel conditions for relaying. By reducing the transmission distance between the SOV and the RSU, this strategy increases the transmission rate and the success rate of model uploads, especially in high-mobility environments. Secondly, energy balancing is achieved by cooperative model transmissions, reducing the power consumption required for direct SOV-to-RSU transmissions. Since received power diminishes rapidly with distance, utilizing OPVs for shorter-range relaying conserves energy, and prevents excessive battery depletion in SOVs. Thirdly, the use of DSTC in cooperative transmissions further enhances upload reliability by leveraging spatial diversity, allowing multiple OPVs to form a virtual multi-antenna system, and enhancing transmission reliability by mitigating channel fading. These mechanisms collectively improve efficiency, reliability, and sustainability in VFL.

\subsection{Complexity Analysis}
The complexity of Algorithm 2 is $\mathcal{O}(T_k|\mathcal{S}_k|(C_d+|\mathcal{U}_k|C_c)),$
where $\mathcal{O}(C_d)$ and $\mathcal{O}(C_c)$ denote the complexity of solving $P3.1$ and $P4$, respectively. $P3.1$ can be solved in constant time $\mathcal{O}(1)$ according to Proposition 1. $P4$ is a convex optimization problem with a convex objective and up to $2|\mathcal{U}_k|+1$ linear constraints, involves an optimization variable of dimension $|\mathcal{U}_k|+1$. Utilizing the interior-point method, $P4$ can be addressed with a complexity of $\mathcal{O}(|\mathcal{U}_k|^{4.5}\ln{\frac{1}{\epsilon}})$ for a given precision $\epsilon$ of the solution. Ignoring lower-order terms, the overall computation complexity of Algorithm 2 is $\mathcal{O}(T_k|\mathcal{S}_k||\mathcal{U}_k|^{5.5}\ln{\frac{1}{\epsilon}}).$

\subsection{Implementation Issues}
To enable the efficient implementation of the VEDS algorithm in a real system, we provide practical considerations as follows.

\subsubsection{Synchronization in VFL Training}
Maintaining synchronization in VFL systems is critical to ensure timely and efficient aggregation of local models from participating vehicles. Time synchronization protocols such as the network time protocol (NTP) can be introduced \cite{syn}, ensuring all devices share a common time reference. NTP works by exchanging time-stamped messages between vehicles and the RSU, measuring the time delay during the transmission and adjusting the local clock of vehicles accordingly. 
\subsubsection{Integration with Existing Vehicular Communication Protocols}
To be implemented in real-world scenarios, the VEDS algorithm needs to be integrated with existing vehicular communication standards such as 5G NR-V2X \cite{V2Vstd2} and LTE-V2X \cite{V2Vstd}. These protocols support both V2I and V2V communications, as well as the exchange of control information, which ensures the algorithm operates effectively within the constraints of current vehicular communication frameworks.

\subsubsection{V2V Communication Overhead}
V2V communications introduce additional overhead, especially in scenarios with high vehicle mobility. Firstly, in dense vehicular networks, network congestion exacerbates communication delays. The VEDS algorithm mitigates this by prioritizing vehicle scheduling according to the amount of model parameters transmitted and channel conditions, to prevent simultaneous transmissions from multiple SOVs within the same area. Secondly, frequent link disconnections occur as vehicles continuously move in and out of the transmission range of each other. Fast connection mechanisms like proactive neighbor discovery \cite{Discovery} can be employed to enable vehicles to quickly detect and connect to new neighbors, reducing signaling overhead during reconnections. Finally, when implemented into practical systems, there might be potential interference from other vehicles not participating in the VFL process (e.g., those transmitting other signals). Our algorithm can be integrated with the sidelink semi-persistent scheduling (SPS) mechanism \cite{SPS1, SPS2}. The SPS mechanism in C-V2X randomly allocates one of the best resources based on channel sensing in the previous period, and keeps transmitting on the selected resource for certain times to minimize the interference and possibility of collisions. Our algorithm can schedule vehicles based on the resources allocated by SPS.

\subsection{Extension to Frequency Selective Channels}
\label{frequencys}
The VEDS algorithm introduced in the previous section assumes frequency non-selective channels. In this section, we extend its deployment to a frequency selective environment, where the additional dimension of channel variation across frequency should be considered. A feasible approach to extending the current design is to utilize the average channel gain for each vehicle, as presented in \cite{comp2, fresele1, fresele2}. For an arbitrary channel $h_{m,n}(t)$, we divide it into $|\mathcal{N}_{m,n}|$ non-selective sub-channels, where $\mathcal{N}_{m,n}$ represents the set of sub-channel indices and $h_{m,n}^{(j)}(t)$ is the channel gain of the $j$-th sub-channel. The average channel gain is then expressed as:
\begin{equation}
\begin{aligned}
    \bar{h}_{m,n}(t) = \frac{\sqrt{\sum_{j \in \mathcal{N}_{m,n}} \left(h_{m,n}^{(j)}(t)\right)^2}}{|\mathcal{N}_{m,n}|}.\notag \\ 
\end{aligned}
\end{equation}
By substituting this average channel gain $\bar{h}_{m,n}(t)$ for $h_{m,n}(t)$, we can apply the VEDS algorithms in a frequency selective context.

\section{Experiments}
In this section, we evaluate the performance of the proposed VEDS algorithm. Firstly, it is compared with the benchmarks under different system parameters. Then, the proposed VEDS algorithm is evaluated for the CIFAR-10 image classification task \cite{cifar10}. Finally, the VEDS algorithm is applied to a real-world trajectory prediction dataset Argoverse \cite{Argoverse} to showcase the value in practical vehicular applications.

\subsection{Simulation Setups}
\begin{figure}[t!]
\centering
\includegraphics[width=0.43\textwidth]{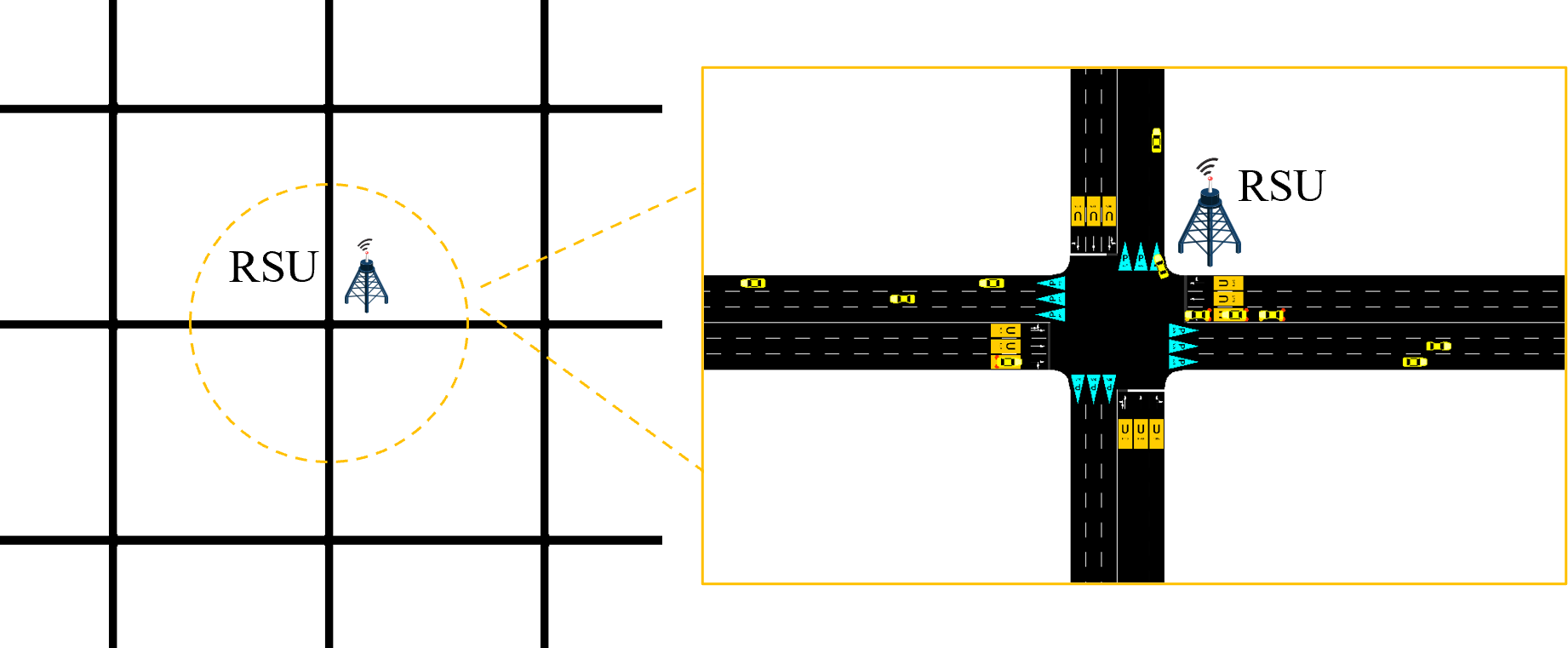}
\caption{The SUMO road network.}
\label{SUMO}
\end{figure}

A road network is built based on SUMO \cite{SUMO}, as shown in Fig. \ref{SUMO}. An RSU is placed at the center of the road network.  The vehicles move according to the Manhattan mobility model. At each intersection, a vehicle has a $50\%$ probability of going straight, and a $25\%$ probability of turning left or right. The vehicle speed is modeled by an intelligent driver model, with a maximum speed of $v^{\text{max}}$ m/s, where $v^{\text{max}}$ is a variable for the experiments. The speed of vehicle $m$ is controlled by the following equation:
$$
a_m=a^{\text{max}}\left[1-\left(\frac{v_m}{v^{\text{max}}}\right)^4-\left(\frac {d_m'} {d_m^{*}}\right)^2\right],
$$
where $v_m$ is the instantaneous speed, $v^{\text{max}}$ is the maximum speed, $a_m$ is the instantaneous acceleration,  $a^{\text{max}}$ is the maximum acceleration, $d_m'$ is the distance between vehicle $m$ and the vehicle in front of it. $d_m^{*}$ is the desired dynamical distance, calculated as
$$
d_m^{*} = d^{\text{safe}} + v_m t^{\text{dst}} + \frac{v_m \cdot \Delta v_m}{2\sqrt{a^{\text{max}} a^{\text{brak}}}}, 
$$
where $d^{\text{safe}}$ is the safety distance between vehicle $m$ and the vehicle in front of it, $t^{\text{dst}}$ is the desired time headway that gives the minimum possible time to the vehicle directly in front of $m$, $\Delta v_m$ is the speed difference between $m$ and the vehicle in front of it, and $a^{\text{brak}}$ is a constant that defines the comfortable braking deceleration.

For wireless communications, we adopt the V2X channel models in 3GPP TR 37.885 \cite{V2Vstd2}. In the urban environment, the pathloss of the LOS and the NLOSv channels are given by $PL_\text{LOS} = 38.77+16.7\log _{10}{d}+18.2\log _{10}{\gamma},$
where $d$ is the distance between two devices, $\gamma$ is the carrier frequency. The pathloss of the NLOS channel is specified by $PL_\text{NLOS} = 36.85+30\log _{10}{d}+18.9\log _{10}{\gamma}.$ The simulation parameters are summarized in Table \ref{table}.

For comparison, the following benchmarks are considered:
\subsubsection{Optimal Benchmark} All the SOVs within the RSU coverage can successfully upload their model parameters.

\subsubsection{Dynamic algorithm with V2I-only communications (V2I-only)} This framework adjusts transmission strategies dynamically in every time slot, considering vehicle mobility. However, it solely uses V2I communications, meaning that the OPVs are not included. This is a special case of our proposed algorithm.

\begin{figure*}[!t]
  \centering
  \begin{minipage}[t]{0.32\textwidth}
    \vspace{0pt} 
    \centering
    \includegraphics[width=\textwidth]{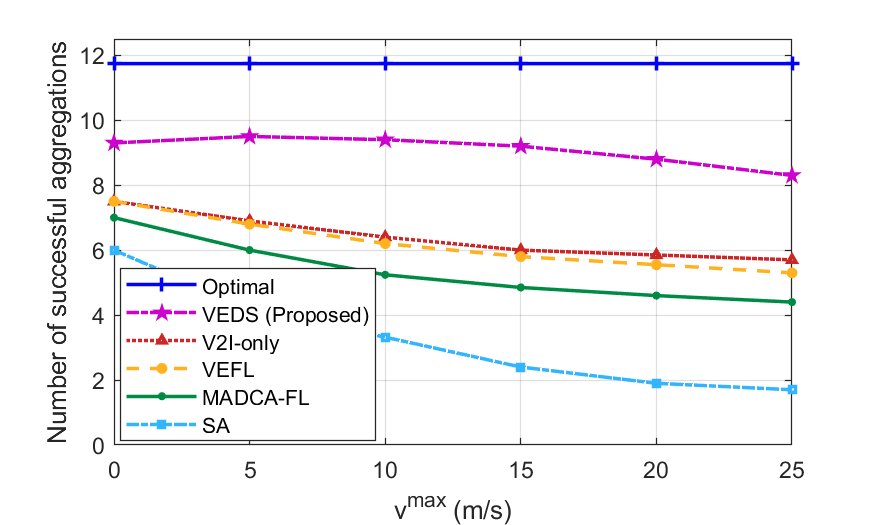}
    \caption{Performance of the VEDS algorithm and the benchmarks under different vehicle speeds.}
    \label{speed}
  \end{minipage}
  \hfill
  \begin{minipage}[t]{0.32\textwidth}
    \vspace{0pt}
    \centering
    \includegraphics[width=\textwidth]{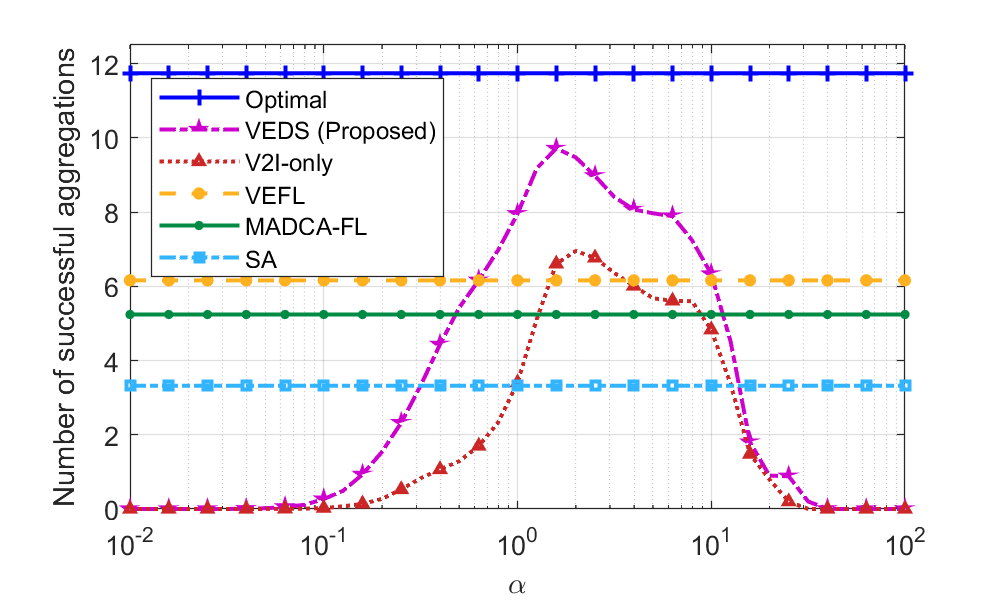}
    \caption{Performance of the VEDS algorithm and the benchmarks under different parameters $\alpha$.}
    \label{alpha}
  \end{minipage}
  \hfill
  \begin{minipage}[t]{0.32\textwidth}
    \vspace{0pt}
    \centering
    \includegraphics[width=\textwidth]{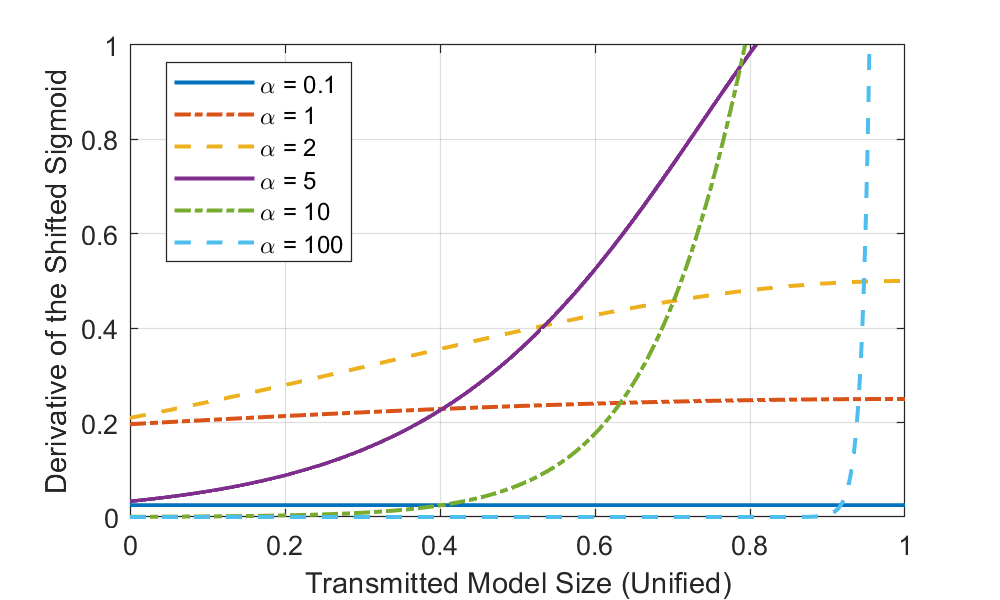}
    \caption{The graph for the derivative of the sigmoid function.}
    \label{dsigmoid}
  \end{minipage}
  \hfill
    \begin{minipage}[t]{0.32\textwidth}
    \vspace{0pt} 
    \centering
    \includegraphics[width=\textwidth]{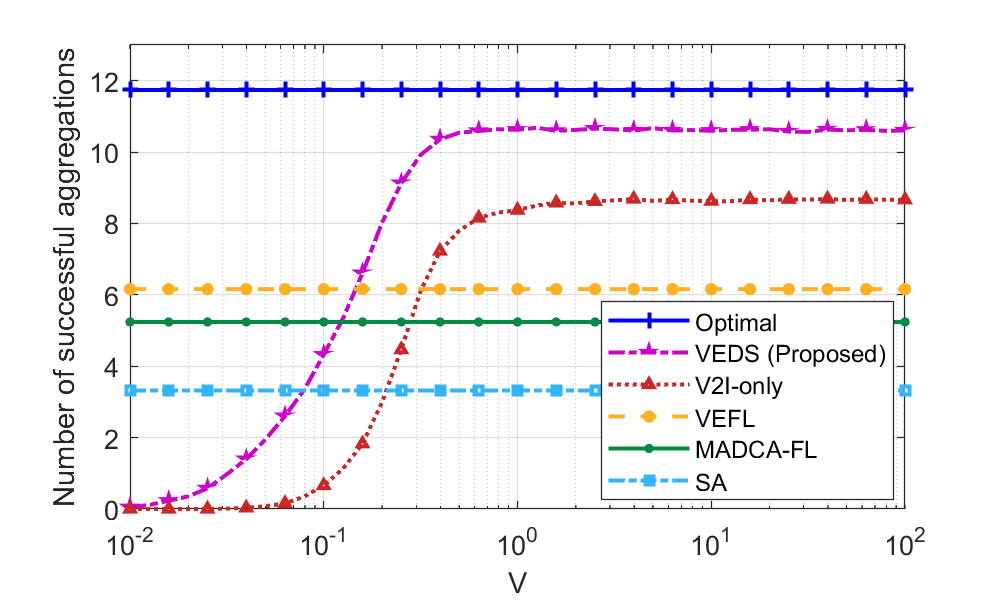}
    \caption{Performance of the VEDS algorithm and the benchmarks under different weight $V$.}
    \label{V1}
  \end{minipage}
  \hfill
  \begin{minipage}[t]{0.32\textwidth}
    \vspace{0pt}
    \centering
    \includegraphics[width=\textwidth]{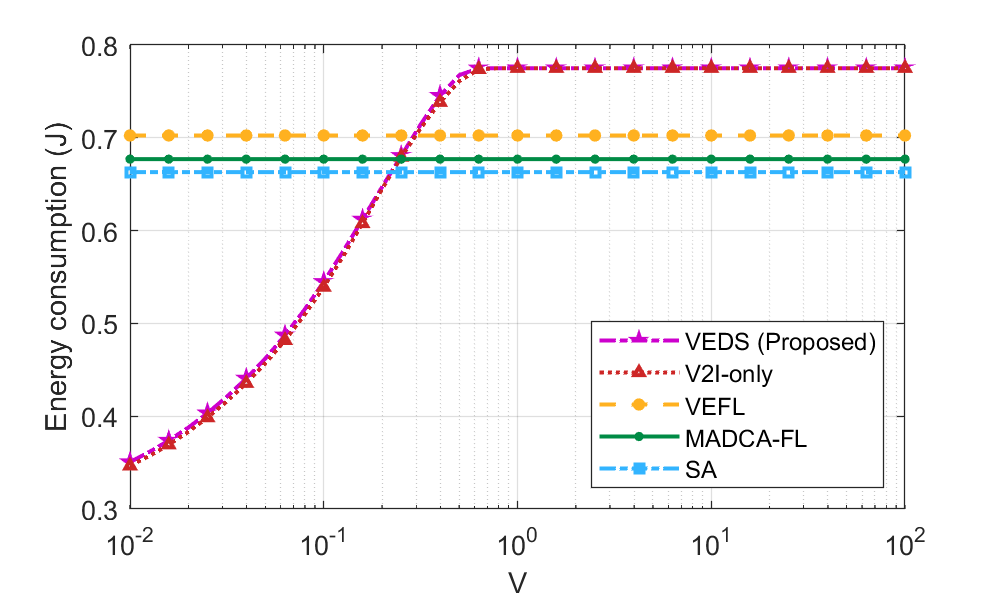}
    \caption{Total energy consumption of all vehicles per round under different weight parameters $V$.}
    \label{V2}
  \end{minipage}
  \hfill
  \begin{minipage}[t]{0.32\textwidth}
    \vspace{0pt}
    \centering
    \includegraphics[width=\textwidth]{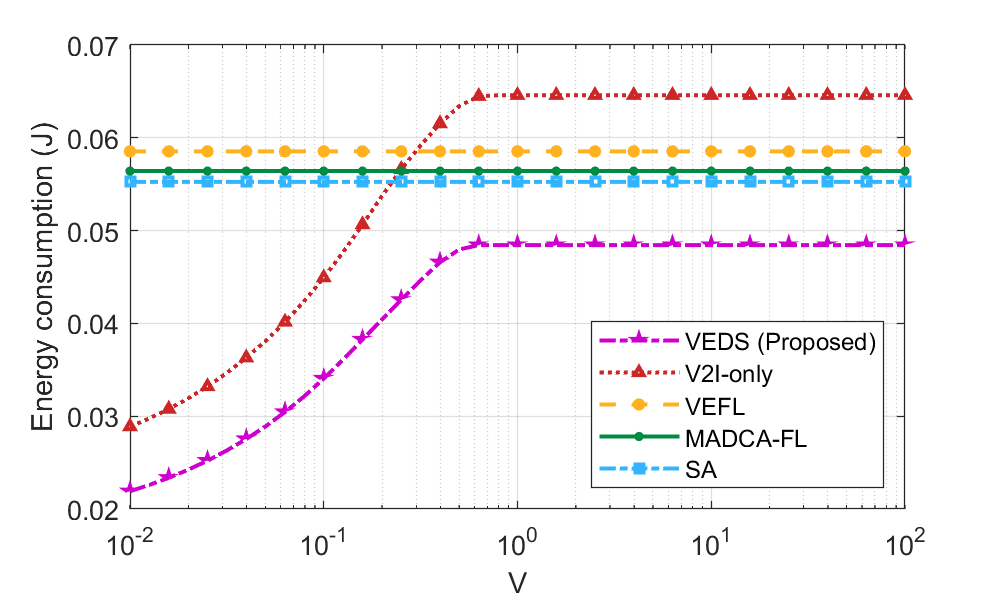}
    \caption{The average energy consumption of SOVs per round under different weight parameters $V$.}
    \label{V3}
  \end{minipage}
\end{figure*}
\subsubsection{Vehicular edge FL (VEFL)} This is a state-of-the-art VFL framework that considers rapidly changing channels and vehicle mobility \cite{VFL8}. This framework considers that the vehicular network operates in a discrete time-slotted manner, and dynamically adjusts the resource allocation decisions in every slot.

\subsubsection{Mobility and channel dynamic-aware FL (MADCA-FL)} This is another state-of-the-art VFL framework considering rapidly changing channels and vehicle mobility \cite{VFL7}. This framework makes scheduling and resource allocation decisions at the start of each round instead of in every slot.

\subsubsection{Static resource allocation and device scheduling algorithm (SA)} This framework does not consider the rapidly time-varying channel and vehicle mobility. It schedules vehicles based on their initial channel states and positions, which is a modified version of the state-of-the-art device scheduling and resource allocation scheme \cite{DS7}.

\subsection{Performance of VEDS under Different Parameters}

\subsubsection{Impact of maximum vehicle speed $v^{\text{max}}$}
Firstly, we validate the performance of our algorithm under different maximum vehicle speeds. We use the objective function of $P1$, i.e., the number of successful aggregations, as the performance metrics. As illustrated in Fig. \ref{speed}, the number of successful aggregations of our framework initially increases and then decreases as $v^{\text{max}}$ is adjusted from 0 (a stationary scenario) to $25$m/s, achieving $81.03\%$ of the optimal benchmark performance when $v^{\text{max}}=5$m/s. 
This performance increase at low speeds can be attributed to the mobility of vehicles allowing OPVs to enter the coverage of the RSU, while the SOVs largely remain within the RSU coverage area. If the vehicles move at high speed, the departure of some SOVs from RSU coverage results in deteriorated channel conditions. However, with the assistance of OPVs, these SOVs can still transmit the model back to the RSU. The comparison of VEDS with the V2I-only framework further highlights the advantages of V2V communications. At low speeds, where most SOVs remain within RSU coverage, VEDS achieves higher performance than V2I-only by dynamically scheduling OPVs with favorable channel conditions for relaying, reducing transmission distance and improving transmission rate. At high speeds, SOVs frequently leave RSU coverage before completing uploads, causing a significant performance drop in V2I-only. VEDS effectively utilizes OPVs to relay models, ensuring stable transmissions despite mobility. Among the other benchmarks, VEFL and the MADCA-FL frameworks also consider vehicle mobility and exhibit certain robustness to changes in mobility. The SA framework, which employs static device scheduling, shows a significant performance decline in high-speed scenarios.

\subsubsection{Impact of $\alpha$} We evaluate our proposed algorithm for different values of $\alpha$, as shown in Fig. \ref{alpha}. It is illustrated that as $\alpha$ increases from $10^{-2}$ to $10^2$, the number of successful aggregations first increases and then decreases, reaching a maximum when $\alpha$ is approximately equal to $2$. This is because as Theorem 3 suggests, when the parameter $\alpha$ is too small, the sigmoid function becomes overly smooth, leading to a suboptimal approximation of the indicator function. On the other hand, when $\alpha$ is too large, the term $\psi(\alpha)$ diminishes, resulting in a loose bound in (\ref{bound}). Both scenarios adversely affect the overall performance of the algorithm. We also explain this phenomenon from a more intuitive perspective. As illustrated in Fig. \ref{dsigmoid}, when $\alpha$ is small, the weight $\frac{d\sigma (\zeta_m(t))}{d\zeta_m(t)}$ increases slowly with respect to $\zeta_m(t)$, the amount of transmitted model parameters. In this case, the algorithm tends to schedule vehicles evenly to balance their energy consumption. Consequently, it is possible that many vehicles have transmitted most of their model parameters but have not completed the upload. In the FL context, such a scenario is considered a transmission failure. When $\alpha$ is large, $\frac{d\sigma (\zeta_m(t))}{d\zeta_m(t)}$ also increases slowly when $\zeta_m(t)$ is small, and thus, the aforementioned phenomenon persists, leading to suboptimal performance.

\begin{table}[t!]
\caption{Simulation Parameters.}
\begin{tabular}{l|l}
\hline
\textbf{Simulation Parameters}                   & \textbf{Values}                            \\ \hline
System bandwidth             & 20MHz                             \\ \hline
Carrier frequency            & 5.9GHz                            \\ \hline
Maximum transmission power   & 0.3W                              \\ \hline
Noise power spectrum density & -174dBm/Hz                        \\ \hline
Shadowing fading std. dev.   & 3dB (LOS, NLOSv), 4dB (NLOS)      \\ \hline
Vehicle blockage loss        & $\max\{0, \mathcal{N}(5, 4)\}$ dB \\ \hline
Energy consumption coefficient      & $10^{-28}$  \cite{energyco1, energyco2}\\ \hline
Energy constraints per vehicles       & Selected from $0.02$J to $0.1$J\\ \hline
Total energy constraints per round     & $0.9$J\\ \hline
Length of time slot & $0.1$s\\ \hline
\end{tabular}
\label{table}
\end{table}
\subsubsection{Impact of $V$}
Then, we evaluate our proposed VEDS algorithm for different weight parameters $V$. The number of successful aggregations and the energy consumption of all vehicles are shown in Fig. \ref{V1} and Fig. \ref{V2}, respectively. It is illustrated that as $V$ increases from  $10^{-2}$ to $10^2$, vehicles tend to consume more energy, resulting in higher energy usage and more successful aggregations. When $V$ exceeds a threshold (around $V=1$), most vehicles use their maximum transmission power to upload their model, and the energy constraints are violated. Therefore, in practical systems, it is crucial to carefully choose the value of $V$ to ensure optimal training performance under energy constraints. Fig. \ref{V3} illustrates the average energy consumption of SOVs per round under different weight parameters $V$. Compared to V2I-only, while the total energy consumption is similar, VEDS achieves better energy balance by distributing transmission tasks among OPVs, reducing the average energy consumption per SOV.


\subsubsection{Impact of Number of SOVs}
Then, we evaluate the scalability of the VEDS algorithm as the number of SOVs in RSU coverage increases, as illustrated in Fig. \ref{density}. When the average number of SOVs in RSU coverage is below $12$, the number of successful aggregations of the proposed VEDS algorithm increases steadily as the number of SOVs grows, achieving close to optimal performance. When the average number of SOVs exceeds $12$, the increasing rate begins to slow down due the the limited communication and energy resources. Despite this, the proposed algorithm outperforms the benchmarks for all numbers of SOVs in the RSU coverage. The V2I-only, VEFL and MADCA-FL frameworks show moderate growth and exhibit performance gaps relative to VEDS. SA performs the worst due to its static scheduling mechanism, which fails to adapt to the increasing number of SOVs. These results demonstrate the scalability of the VEDS algorithm in accommodating a growing number of SOVs.

\subsubsection{Impact of Model Size}
\begin{figure*}[!t]
  \centering
  \begin{minipage}[t]{0.32\textwidth}
    \vspace{0pt} 
    \centering
    \includegraphics[width=\textwidth]{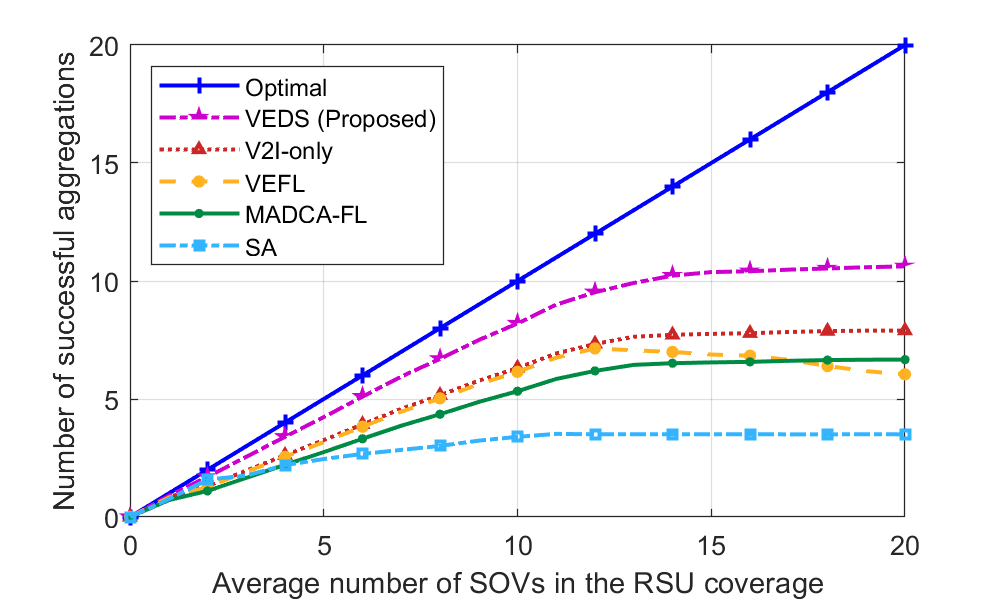}
    \caption{Performance of the VEDS algorithm and the benchmarks under different numbers of SOVs.}
    \label{density}
  \end{minipage}
  \hfill
  \begin{minipage}[t]{0.32\textwidth}
    \vspace{0pt}
    \centering
    \includegraphics[width=\textwidth]{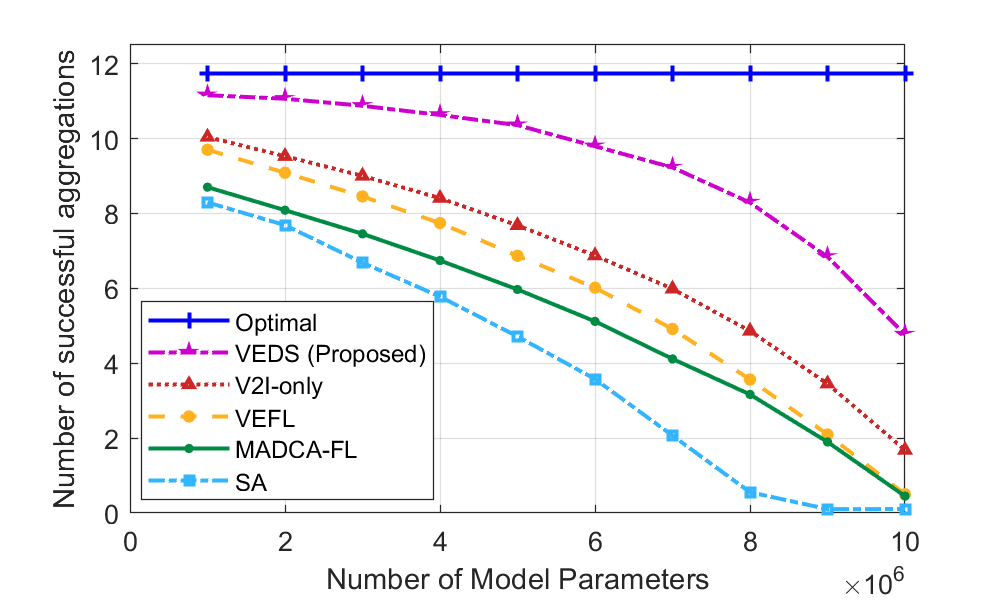}
    \caption{Performance of the VEDS algorithm and the benchmarks under different model sizes.}
    \label{modelsize}
  \end{minipage}
  \hfill
  \begin{minipage}[t]{0.32\textwidth}
    \vspace{0pt} 
    \centering
    \includegraphics[width=\textwidth]{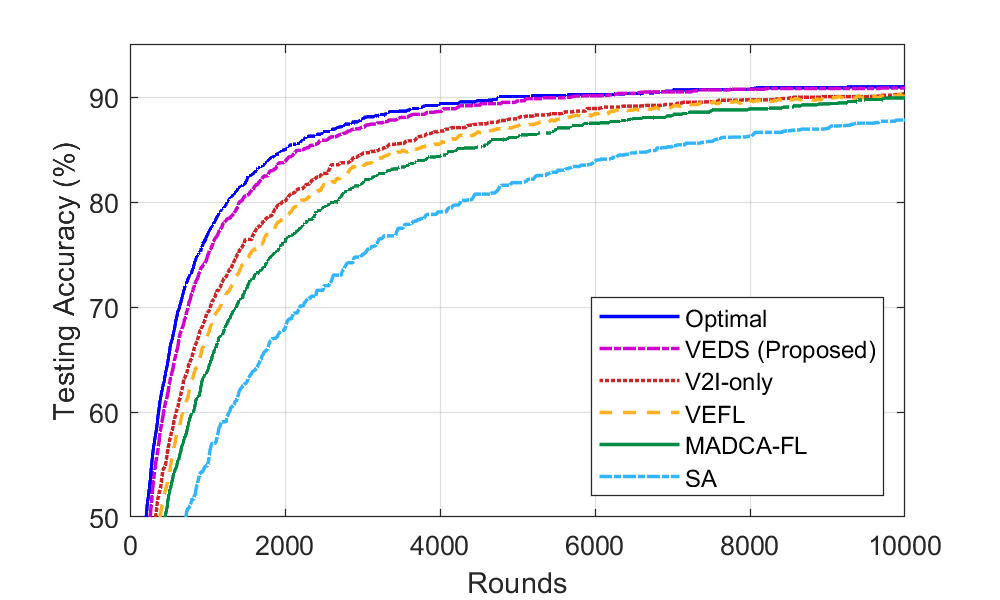}
    \caption{Test accuracy on the CIFAR-10 dataset (\emph{i.i.d.} setting).}
    \label{cifariid}
  \end{minipage}
  \hfill
  \begin{minipage}[t]{0.32\textwidth}
    \vspace{0pt}
    \centering
    \includegraphics[width=\textwidth]{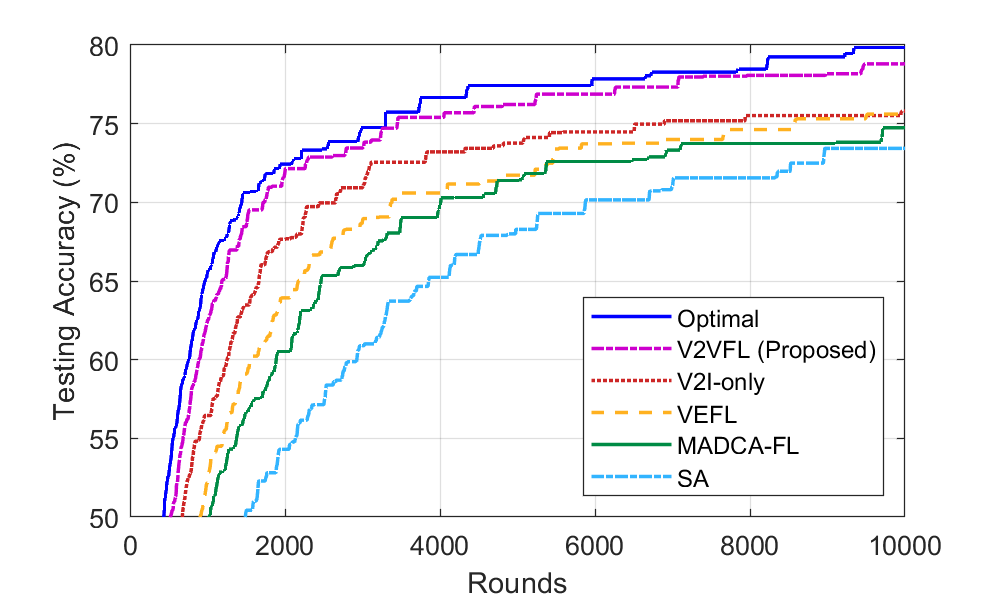}
    \caption{Test accuracy on the CIFAR-10 dataset (\emph{non-i.i.d.} setting).}
    \label{cifarniid}
  \end{minipage}
    \hfill
  \begin{minipage}[t]{0.32\textwidth}
    \vspace{0pt}
    \centering
    \includegraphics[width=\textwidth]{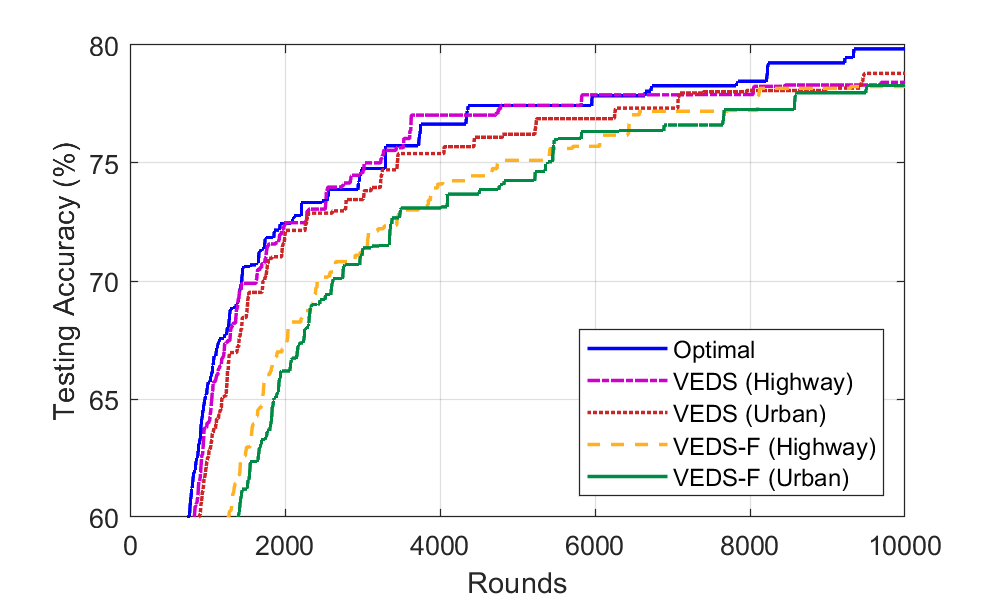}
    \caption{Test accuracy on the CIFAR-10 dataset compared with frequency selective channel and highway scenarios (\emph{non-i.i.d.} setting).}
    \label{cifarfreq}
  \end{minipage}
  \begin{minipage}[t]{0.32\textwidth}
    \vspace{0pt} 
    \centering
    \includegraphics[width=\textwidth]{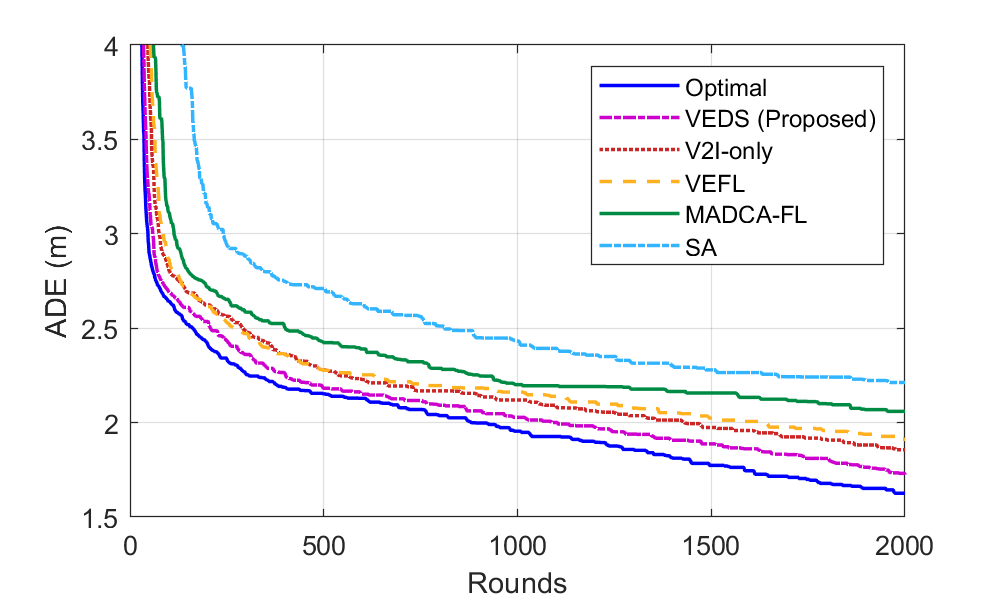}
    \caption{ADE for the trajectory prediction task on the Argoverse dataset.}
    \label{adeiid}
  \end{minipage}
  \hfill
\end{figure*}

We evaluate the performance of our algorithm under different model sizes, as shown in Fig. \ref{modelsize}. For smaller model sizes (e.g., 1 or 2 million parameters), the number of successful aggregations of VEDS is close to the optimal value. As the model size increases, the number of successful aggregations gradually decreases due to the growing communication overhead. In contrast, other benchmarks show a sharper performance decline as the model size increases, particularly under larger model sizes, indicating less adaptability to increased model size. 
\subsection{Evaluation on the CIFAR-10 Dataset}
Then, we evaluate the proposed VEDS algorithm on the CIFAR-10 dataset\cite{cifar10}, which comprises $50000$ training images and $10000$ test images across ten categories. We consider both the independent and identically distributed (\emph{i.i.d.}) and non-independent and identically distributed (\emph{non-i.i.d}) settings. For the \emph{i.i.d.} setting, the dataset is evenly divided into $40$ subsets, each containing samples from all 10 categories. For the \emph{non-i.i.d}. setting, data samples are organized by category, and each vehicle holds a disjoint subset of data with samples from $2$ categories.  Using the dataset, we train a convolutional neural network (CNN). This CNN includes six convolutional layers. Three of them are followed by a ReLU activation layer and a max pooling layer, and the other three are followed by a ReLU activation layer and a normalization layer. The last convolutional layer is connected to a fully connected layer and a softmax output layer for classification. The batch size of each vehicle is randomly chosen from 16, 32 and 48, and the learning rate is set to 0.1. 

The test accuracy of the VEDS algorithm compared with the benchmarks is illustrated in Fig. \ref{cifariid} (\emph{i.i.d.}) and Fig. \ref{cifarniid} (\emph{non-i.i.d.}), where $\alpha =2, V=0.2$, and the maximum vehicle speed $v^{\text{max}}=10$ m/s. In the \emph{i.i.d.} scenario, both VEDS and the benchmarks achieve high test accuracy. 
The VFL convergence speed of the VEDS algorithm closely approaches that of the optimal benchmark and is significantly higher than other benchmarks. Under the \emph{non-i.i.d.} scenario, the convergence speed and the highest test accuracy of the VEDS algorithm are close to the optimal benchmark and exceed the other three benchmarks. After $1000$ rounds of training, VEDS achieves a test accuracy of $62.42\%$, exceeding V2I-only, VEFL, MADCA-FL and SA over $10.57\%$, $19.41\%$  $26.90\%$ and $41.42\%$. After $10000$ rounds of training, the highest achievable accuracies are $79.84\%$ for the optimal benchmark, $78.78\%$ for VEDS, $75.80\%$ for V2I-only, $75.61\%$ for VEFL, $74.74\%$, and $73.23\%$ for SA.

We also evaluate the performance of the VEDS algorithm under frequency selective channels (denoted as VEDS-F), as discussed in Section \ref{frequencys}. Additionally, we extend the evaluation to highway scenarios, where the pathloss models for LOS and NLOSv channels are modified by $PL_\text{LOS} = 38.77+16.7\log _{10}{d}+18.2\log _{10}{\gamma}$. Fig. \ref{cifarfreq} shows that the highest testing accuracy in highway environments is slightly better, exceeding that in urban environments by $0.16\%$. For the frequency selective channel scenario, we use the average channel condition to estimate the actual channel state, which introduces a slight performance degradation. While the convergence speed of the VEDS algorithm under frequency selective fading channels is slightly lower than that under frequency non-selective fading channels, the highest testing accuracy is well-preserved. The highest testing accuracy under the frequency selective fading scenario reaches $78.44\%$ for the highway case and $78.28\%$ for the urban case, which is $0.60\%$ and $0.64\%$ lower than the frequency non-selective fading scenario, respectively. These results demonstrate the applicability of the VEDS algorithm under various vehicular environments and channel conditions.


\subsection{Evaluation on Argoverse Trajectory Prediction Dataset}
Finally, we evaluate the proposed VEDS algorithm on the real-world trajectory prediction dataset Argoverse \cite{Argoverse}. Argoverse encompasses more than $300000$ sequences gathered from Pittsburgh and Miami. Each sequence is captured from a moving vehicle at a sampling frequency of $10$ Hz. The task is to predict the position of the vehicle for the next 3 seconds. The dataset is organized into training, validation, and test sets, containing $205942$, $39472$, and $78143$ sequences, respectively. The sequences are uniformly partitioned into 40 subsets.

 Based on the dataset, the VFL system collaboratively trains a lane graph convolutional neural network (LaneGCN) \cite{lane}. The LaneGCN includes three sub-neural networks: an ActorNet, a MapNet, and a FusionNet. The ActorNet contains a 1D CNN and a Feature Pyramid Network (FPN) to extract features of vehicle trajectories. The MapNet is a graph convolutional neural network that represents and extracts the map features. The FusionNet is used to fuse the vehicle trajectory features and the map features to output the final trajectory prediction results. The batch size of each vehicle is randomly chosen from 16, 32 and 48, and the learning rate is set to 0.1. We employ ADE as the metric for trajectory prediction, which is the average $l_2$ distance between the actual and predicted vehicle positions on the trajectory.

The performance of the proposed framework compared with the benchmarks is illustrated in Fig. \ref{adeiid}. It is shown that the proposed VEDS algorithm outperforms the benchmarks both in terms of ADE. Specifically, VEDS achieves an ADE of $1.72$ after $2000$ rounds of training, which is $7.14\%$, $9.82\%$, $16.30\%$ and $22.09\%$ lower than V2I-only, VEFL, MADCA-FL and SA, respectively. These results validate the strong performance of our proposed VEDS algorithm when applied to real-world autonomous driving datasets.

\section{Conclusions}
This paper has considered a VFL system, where the SOVs and OPVs in a vehicular network collaborate to train an ML model under the orchestration of the RSU. A VEDS algorithm has been proposed to optimize the VFL training performance under energy constraints and channel uncertainty of vehicles. A convergence analysis has been performed to transform the implicit FL loss function into the number of successful aggregations. Then, a derivative-based drift-plus-penalty method has been proposed to convert the long-term stochastic optimization problem into an online MINLP problem, and a theoretical performance guarantee has been provided to bound the performance gap between the online and offline solutions. Based on the analysis of the scheduling priority, the MINLP problem has been further reduced to a set of convex optimization problems, solved by the interior-point method. Experimental results have demonstrated the robustness of the framework under varying vehicle speeds. The test accuracy is increased by $4.20\%$ for the CIFAR-10 dataset, and the ADE is reduced by $9.82\%$ for the Argoverse dataset.

In the future, we plan to extend our current framework by incorporating sparsification and quantization to reduce the communication overhead. Furthermore, we aim to develop scheduling strategies that consider the diverse distributions of local datasets across different to increase the model robustness and generalization. We will also explore the application of our proposed method on a broader range of datasets, including tasks such as autonomous driving object detection and semantic segmentation. These future developments will address current limitations and expand the practical applicability of our framework to complex and diverse scenarios.

\appendices

\section{Proof of Lemma 1}
\label{lemma1}
For simplicity, we use $I_{m,k}$ to denote $\mathbb{I}\left(\sum_{t\in \mathcal{T}_k} z_m(t) \geq Q\right)$ in the appendix. According to Assumption 1 and definition (\ref{global}), the global loss function is also $L$-smooth and $\mu$-strongly convex. There is:
\begin{equation}
\begin{aligned}F&(\boldsymbol {w}_{k}) - F(\boldsymbol {w}_{k-1})  \\
&\leq \left < { \nabla F(\boldsymbol {w}_{k-1}), \boldsymbol {w}_{k}-\boldsymbol {w}_{k-1} }\right >  + \frac {L}{2} \left \Vert{ \boldsymbol {w}_{k}-\boldsymbol {w}_{k-1} }\right \Vert ^{2}.\label{conv1}
\end{aligned}
\end{equation}
For the term $\left < { \nabla F(\boldsymbol {w}_{k-1}), \boldsymbol {w}_{k+1}-\boldsymbol {w}_{k} }\right >$, we have 
\begin{subequations}
\begin{align}
& \mathbb{E}\left[\left < { \nabla F(\boldsymbol {w}_{k-1}), \boldsymbol {w}_{k}-\boldsymbol {w}_{k-1} }\right >\right]\notag \\ 
&=   \mathbb{E}\left[ \big < { \nabla F(\boldsymbol {w}_{k-1}), \frac{\sum_{m\in {\mathcal{S}}_k} I_{m,k} (\boldsymbol {w}_{m,k}-\boldsymbol {w}_{k-1})}{\sum_{m\in \mathcal{S}_k} I_{m,k} }} \big >\right]\notag\\
&=   -\mathbb{E}\left[ \big < { \nabla F(\boldsymbol {w}_{k-1}),\frac{\sum_{m\in {\mathcal{S}}_k} I_{m,k} \eta_k \nabla f\left(\boldsymbol{w}_{k-1},\mathcal{B}_{m,k}\right)}{\sum_{m\in \mathcal{S}_k}  I_{m,k} }} \big >\right]\notag\\
&= -\eta_k  \left \Vert{ \nabla F(\boldsymbol {w}_{k-1}) }\right \Vert ^2.\tag{32} \label{term1}
\end{align}
\end{subequations}
For the term $\left \Vert{ \boldsymbol {w}_{k}-\boldsymbol {w}_{k-1} }\right \Vert^2$, we have 
\begin{subequations}
\begin{align}
& \mathbb{E}\left[ \left \Vert{ \boldsymbol {w}_{k}-\boldsymbol {w}_{k-1} }\right \Vert ^{2} \right]\notag\\ 
&= \mathbb{E}\left[ \left \Vert{ \frac{\sum_{m\in {\mathcal{S}}_k} I_{m,k} |\mathcal{D}_m|(\boldsymbol {w}_{m,k}-\boldsymbol {w}_{k-1})}{\sum_{m\in \mathcal{S}_k} I_{m,k} |\mathcal{D}_m|}  }\right \Vert ^{2} \right]\notag\\
&=   \mathbb{E}\left[ \left \Vert{ \frac{\sum_{m\in {\mathcal{S}}_k} \sum_{\boldsymbol{x}\in\mathcal{B}_{m,k}} I_{m,k} |\mathcal{D}_m|\eta_k \nabla f\left(\boldsymbol{w}_{k-1};\boldsymbol{x}\right)}{B_{k}\sum_{m\in \mathcal{S}_k}  I_{m,k} |\mathcal{D}_m|} }\right \Vert ^{2} \right]\notag\\
&\leq \eta_k^2 \left \Vert{ \nabla F(\boldsymbol {w}_{k-1}) }\right \Vert^2 + \frac{\eta_k^2 G^2}{B_{k}\sum_{m\in \mathcal{S}_k}  I_{m,k}}. \tag{33} \label{term2}
\end{align}
\end{subequations}
Taking the expectation over stochastic data sampling on both sides of (\ref{conv1}) and plugging (\ref{term1}) and (\ref{term2}), we have
\begin{equation}
\begin{aligned}
&\mathbb{E}[F(\boldsymbol{w}_k)]-\mathbb{E}[F(\boldsymbol{w}_{k-1})]  \leq  -\eta_k  \left \Vert{ \nabla F(\boldsymbol {w}_{k-1}) }\right \Vert ^2 \\ &+  \frac{L\eta_k^2}2 \left( \left \Vert{ \nabla F(\boldsymbol {w}_{k-1}) }\right \Vert^2 +\frac{G^2}{B_{k}\sum_{m\in \mathcal{S}_k} I_{m,k}} \right)\\
&= \eta_k\left(\frac{L\eta_k}2-1\right) \left \Vert{ \nabla F(\boldsymbol {w}_{k-1}) }\right \Vert^2 +\frac{L\eta_k^2}2 \frac{G^2}{B_{k}\sum_{m\in \mathcal{S}_k} I_{m,k}}.\notag
\end{aligned}
\end{equation}
Lemma 1 is proved.
\section{Proof of Theorem 1}
\label{Theorem1}
Assuming that the loss function is the loss $\mu$-strongly convexity, the Polyak-Lojasiewicz inequality holds 
\begin{equation} \lVert \nabla F(\boldsymbol {w}_{k-1}) \rVert^{2} \geq 2\mu (F({\boldsymbol {w}}_{k-1})-F(\boldsymbol {w}^*)).\label{T1E1}\end{equation}
Substituting (\ref{T1E1}) into (\ref{ubound}), and set $\eta_k \leq \frac{1}{L}$, there is
\begin{equation}
\begin{aligned}
&\mathbb {E}[F(\boldsymbol {w}_{k})]-\mathbb{E}[F(\boldsymbol {w}_{k-1})] \\&\leq\eta_k\left(\frac{L\eta_k}2-1\right) \left \Vert{ \nabla F(\boldsymbol {w}_{k-1}) }\right \Vert^2 +\frac{L\eta_k^2}2 \frac{G^2}{B_{k}\sum_{m\in \mathcal{S}_k} I_{m,k}}\\&\leq-\eta _{t} \mu (\mathbb {E}[F({\boldsymbol {w}}_{k-1})]-F(\boldsymbol {w}^*)) +\frac {\eta _{k}}{2}\frac{G^2}{B_{k}\sum_{m\in \mathcal{S}_k} I_{m,k}}.\notag
\end{aligned}
\end{equation}
With recursion, there is
\begin{equation}
\begin{aligned}
&\mathbb {E}[F({\boldsymbol {w}}_{K})]-F(\boldsymbol {w}^*)\leq(1-\mu \eta _{K}) (\mathbb {E}[F({\boldsymbol {w}}_{K-1})]-F(\boldsymbol {w}^*))\\
&  +\frac {\eta _{K}}{2}\frac{G^2}{B_{K}\sum_{m\in \mathcal{S}_K} I_{m,K}} \\&\leq\cdots \leq (\mathbb {E}[F({\boldsymbol {w}}_{0})]-F(\boldsymbol {w}^*))\prod _{k=1}^{K} (1-\mu \eta _{k}) \\&+\,\sum _{k=1}^{K-1} \frac {\eta _{k}}{2}\frac{G^2}{B_{k}\sum_{m\in \mathcal{S}_k} I_{m,k}} \prod _{j=k+1}^{K}(1-\mu \eta _{j}) \\
&+\frac {\eta _{K}}{2}\frac{G^2}{B_{K}\sum_{m\in \mathcal{S}_k} I_{m,K}}.\notag
\end{aligned}\end{equation}
Theorem 1 is proved.

\section{Proof of Theorem 2}
\label{Theorem2}
Based on lemma 1, and setting $\eta_k \leq \frac{1}{L}$, we have
\begin{equation}
\begin{aligned}&\frac {\eta_k} {2}\mathbb{E}\left\| \nabla F(\boldsymbol {w}_{k-1})\right\|^2 \\& \leq \mathbb{E}[F(\boldsymbol{w}_{k-1})]-\mathbb{E}[F(\boldsymbol{w}_{k})]  + \frac{L\eta_k^2}2\frac{G^2}{B_{k}\sum_{m\in \mathcal{S}_k} I_{m,k}}.
\notag
\end{aligned}
\end{equation}

Taking a telescopic sum from $k = 1$ to $k = K$, and setting $\eta_k = \frac{1}{K^{1/2} L} $ we get
\begin{equation}
\begin{aligned}
&\frac {1} {K}\sum_{k=1}^{K} \mathbb{E}\left\|\nabla F(\boldsymbol {w}_{k-1})\right\|^2 \\
&\leq \frac{2L\left(F(\boldsymbol{w}_{0})-\mathbb{E}[F(\boldsymbol{w}_{K})]\right)}{K^{1/2}} + \frac {1} {K^{3/2}}\sum_{k=1}^{K}  \frac{G^2}{B_{k}\sum_{m\in \mathcal{S}_k} I_{m,k}}\\
&\leq \frac{2L\left(F(\boldsymbol{w}_{0})- F(\boldsymbol{w}^{*})\right)}{K^{1/2}} + \frac {1} {K^{3/2}}\sum_{k=1}^{K}  \frac{G^2}{B_{k}\sum_{m\in \mathcal{S}_k} I_{m,k}}.\notag
\end{aligned}
\end{equation}
This completes the proof.

\section{Proof of Theorem 3}
\label{Theorem3}
We define a quadratic Lyapunov function as 
$$L(t) \triangleq\frac{1}{2} \sum_{m\in \mathcal{S}_k}q_m^{\text{SOV}} (t)^2+\frac{1}{2}\sum_{n\in \mathcal{U}_k}q_{n}^{\text{OPV}}(t)^2.$$ 
We define $\delta_m^{\text{SOV}}(t) \triangleq e^{\text{cm}}_m(t)-\frac{E^{\text{cons}}_m - e^{\text{cp}}_{m,k}}{T_k}$, $\delta_n^{\text{OPV}}(t) \triangleq e^{\text{cm}}_n(t)-\frac{E^{\text{cons}}_n}{T_k}$, $\phi_m^{\text{SOV}} \triangleq \max_t\{|\delta_m^{\text{SOV}}(t)|\}$, $\phi_n^{\text{OPV}} \triangleq \max_t\{|\delta_n^{\text{OPV}}(t)|\}$, and $\Phi \triangleq \sum_{m\in \mathcal{S}_k}(\phi_m^{\text{SOV}})^2+\sum_{n\in \mathcal{U}_k}(\phi_n^{\text{OPV}})^2$. 
Then, the Lyapunov drift of a single round is defined as
\begin{subequations}
\begin{align}
    \Delta(t)&\triangleq L(t+1)-L(t)\notag\\& = \frac{1}{2} \sum_{m\in \mathcal{S}_k}\left(q_{m}^{\text{SOV}}(t+1)^2-q_{m}^{\text{SOV}}(t)^2\right)\notag\\ &+  \frac{1}{2} \sum_{n\in \mathcal{U}_k}\left(q_{n}^{\text{OPV}}(t+1)^2-q_{n}^{\text{OPV}}(t)^2\right)\notag\\ 
    &\leq\frac{1}{2} \sum_{m\in \mathcal{S}_k}\left(\left(q_{m}^{\text{SOV}}(t)+\delta_m^{\text{SOV}}(t)\right)^2-q_{m}^{\text{SOV}}(t)^2\right)\ \label{lypdrift}\tag{35}
    \\&+\frac{1}{2} \sum_{n\in \mathcal{U}_k}\left(\left(q_{n}^{\text{OPV}}(t)+\delta_n^{\text{OPV}}(t)\right)^2-q_{n}^{\text{OPV}}(t)^2\right)\notag\\ 
    &\leq \Phi+\sum_{m\in \mathcal{S}_k}q_{m}^{\text{SOV}}(t)\delta_m^{\text{SOV}}(t)+\sum_{n\in \mathcal{U}_k}q_{n}^{\text{OPV}}(t)\delta_n^{\text{OPV}}(t),\notag
\end{align}
\end{subequations}
By adding $-V \sum_{m\in \mathcal{S}_k}z_m (t) \frac{d\sigma (\zeta_m(t))}{d\zeta_m(t)}$ on both sides of (\ref{lypdrift}), the upper bound on the derivative-based drift-plus-penalty function is
\begin{equation}
\begin{aligned}
&\Delta(t)-V \sum_{m\in \mathcal{S}_k}z_m (t) \frac{d\sigma (\zeta_m(t))}{d\zeta_m(t)}\leq \Phi+\sum_{m\in \mathcal{S}_k}q_{m}^{\text{SOV}}(t)\delta_m^{\text{SOV}}(t)\\ 
&+\sum_{n\in \mathcal{U}_k}q_{n}^{\text{OPV}}(t)\delta_n^{\text{OPV}}(t)-V \sum_{m\in \mathcal{S}_k} z_m (t) \frac{d\sigma (\zeta_m(t))}{d\zeta_m(t)}.\notag
\end{aligned}
\end{equation}
We define the $T_k$-round drift as \begin{equation}
\begin{aligned}\Delta_{T_k}&\triangleq  \Delta(T_k+1)-\Delta(1) \\&= \sum_{m\in \mathcal{S}_k}\frac{1}{2}q_{m}^{\text{SOV}}(T_k+1)^2+\sum_{n\in \mathcal{U}_k}\frac{1}{2}q_{n}^{\text{OPV}}(T_k+1)^2.\notag\end{aligned}
\end{equation} Then, the $T_k$-round drift-plus-penalty function is bounded by:
\begin{subequations}
\begin{align}
&\Delta_{T_k}-V \sum_{t\in \mathcal{T}_k} \sum_{m\in \mathcal{S}_k}z_m^\dagger (t) \frac{d\sigma (\zeta_m(t))}{d\zeta_m(t)}\notag\\ 
&\leq T_k \Phi-V\sum_{t\in \mathcal{T}_k} \sum_{m\in \mathcal{S}_k} z_m^\dagger (t) \frac{d\sigma (\zeta_m(t))}{d\zeta_m(t)} \notag\\&+\sum_{t\in \mathcal{T}_k} \left(\sum_{m\in \mathcal{S}_k}q_{m}^{\text{SOV}}(t)\delta_m^{\text{SOV}\dagger}(t)+\sum_{n\in \mathcal{U}_k}q_{n}^{\text{OPV}}(t)\delta_n^{\text{OPV}\dagger}(t)\right)\notag\\
&\overset{(a)}{\leq}T_k \Phi-V\sum_{t\in \mathcal{T}_k} \sum_{m\in \mathcal{S}_k} z_m^* (t) \frac{d\sigma (\zeta_m(t))}{d\zeta_m(t)} \tag{36}\label{lyp1} \\&+\sum_{t\in \mathcal{T}_k} \left(\sum_{m\in \mathcal{S}_k}q_{m}^{\text{SOV}}(t)\delta_m^{\text{SOV}*}(t)+\sum_{n\in \mathcal{U}_k}q_{n}^{\text{OPV}}(t)\delta_n^{\text{OPV}*}(t)\right)\notag,
\end{align}
\end{subequations}
where inequality $(a)$ holds because solving $P3$ yields a minimum value of (\ref{objp3}).

Based on the definition of $q_{m}^{\text{SOV}}(t)$, we have $q_{m}^{\text{SOV}}(t+1) - q_{m}^{\text{SOV}}(t)\leq \phi_m^{\text{SOV}}, \forall m \in \mathcal{S}_k,t \in \mathcal{T}_k$, and therefore
\begin{equation}
\begin{aligned}
&q_{m}^{\text{SOV}}(t) = q_{m}^{\text{SOV}}(t)-q_{m}^{\text{SOV}}(1) \\
&= \sum_{\tau=1}^{t-1}\left( q_{m}^{\text{SOV}}(t+1)-q_{m}^{\text{SOV}}(t) \right) \leq (t-1) \phi_m^{\text{SOV}},\quad  \forall m\in\mathcal{S}_k\notag,
\end{aligned}
\end{equation}
and
\begin{equation}
q_{m}^{\text{SOV}}(t)\delta_m^{\text{SOV}*}(t) \leq (t-1)(\phi_m^{\text{SOV}})^2,\quad \forall m\in\mathcal{S}_k.\label{lyp2}
\end{equation}
Similarly, there is
\begin{equation}
q_{n}^{\text{OPV}}(t)\delta_n^{\text{OPV}*}(t) \leq (t-1)(\phi_n^{\text{OPV}})^2,\quad \forall n\in\mathcal{U}_k.\label{lyp3}
\end{equation}
Substituting (\ref{lyp2}) and (\ref{lyp3}) into (\ref{lyp1}), we have
\begin{equation}
\begin{aligned}
&\Delta_{T_k}-V \sum_{t\in \mathcal{T}_k} \sum_{m\in \mathcal{S}_k}z_m^\dagger (t) \frac{d\sigma (\zeta_m(t))}{d\zeta_m(t)}\\ 
&\leq T_k \Phi-V \sum_{t\in \mathcal{T}_k} \sum_{m\in \mathcal{S}_k} z_m^* (t) \frac{d\sigma (\zeta_m(t))}{d\zeta_m(t)} \\& 
+\sum_{t\in \mathcal{T}_k} \left( \sum_{m\in \mathcal{S}_k}(t-1)(\phi_m^{\text{SOV}})^2+\sum_{n\in \mathcal{U}_k}(t-1)(\phi_n^{\text{OPV}})^2\right)\\
&=T_k^2 \Phi - V  \sum_{t\in \mathcal{T}_k} \sum_{m\in \mathcal{S}_k} z_m^* (t) \frac{d\sigma (\zeta_m(t))}{d\zeta_m(t)}.\notag
\end{aligned}
\end{equation}
Since $\Delta_{T_k}\geq 0$, we have
\begin{equation}
\begin{aligned}
\sum_{t\in \mathcal{T}_k} \sum_{m\in \mathcal{S}_k}& z_m^\dagger (t) \frac{d\sigma (\zeta_m(t))}{d\zeta_m(t)} \\ 
&\geq    \sum_{t\in \mathcal{T}_k} \sum_{m\in \mathcal{S}_k} z_m^* (t) \frac{d\sigma (\zeta_m(t))}{d\zeta_m(t)} - \frac{T_k^2 \Phi}{V}.\notag
\end{aligned}
\end{equation}
Since the function $\sigma (\cdot)$ is continuous and derivable, there exist a point $\xi_m \in (\sum_{t\in \mathcal{T}_k} z_m^\dagger(t), \sum_{t\in \mathcal{T}_k} z_m^*(t))$ such that
\begin{equation}
\frac{\partial \sigma (\xi_m)}{\partial \zeta_m(t)} = \frac{ \sigma\left(\sum_{t\in \mathcal{T}_k} z_m^*(t)\right) -  \sigma\left(\sum_{t\in \mathcal{T}_k} z_m^\dagger(t)\right)}{ \sum_{t\in \mathcal{T}_k} z_m^* (t) -  \sum_{t\in \mathcal{T}_k} z_m^\dagger (t)}.\notag
\end{equation}
Based on (\ref{psi}), we have
\begin{equation}
\begin{aligned}
&\frac{T_k^2 \Phi}{V} \geq \sum_{t\in \mathcal{T}_k} \sum_{m\in \mathcal{S}_k} \left( z_m^* (t) - z_m^\dagger (t) \right) \frac{d\sigma (\zeta_m(t))}{d\zeta_m(t)}\\
&\geq \sum_{m\in \mathcal{S}_k} \left( \sum_{t\in \mathcal{T}_k} z_m^* (t) -  \sum_{t\in \mathcal{T}_k} z_m^\dagger (t) \right) \psi(\alpha) \frac{\partial \sigma (Q)}{\partial \zeta_m(t)}\\
&\geq \sum_{m\in \mathcal{S}_k} \left( \sum_{t\in \mathcal{T}_k} z_m^* (t) -  \sum_{t\in \mathcal{T}_k} z_m^\dagger (t) \right) \psi(\alpha) \frac{\partial \sigma (\xi_m)}{\partial\zeta_m(t)}\\
&= \psi(\alpha) \left[ \sum_{m\in \mathcal{S}_k} \sigma\left(\sum_{t\in \mathcal{T}_k} z_m^*(t)\right) - \sum_{m\in \mathcal{S}_k} \sigma\left(\sum_{t\in \mathcal{T}_k} z_m^\dagger(t)\right)\right].\notag
\end{aligned}
\end{equation}
Finally, there is
\begin{equation}
\sum_{m\in \mathcal{S}_k} \sigma\left(\sum_{t\in \mathcal{T}_k} z_m^*(t)\right) - \sum_{m\in \mathcal{S}_k} \sigma\left(\sum_{t\in \mathcal{T}_k} z_m^\dagger(t)\right) \leq  \frac{T_k^2 \Phi}{V \psi(\alpha)}.\notag
\end{equation}
For energy consumption, we have
\begin{equation}
\begin{aligned}
&\sum_{t\in\mathcal{T}_k} \left( e^{\text{cm}}_m(t)-\frac{E^{\text{cons}}_m}{T_k} \right) +e^{\text{cp}}_{m,k} \leq \sum_{t\in\mathcal{T}_k} q_{m}^{\text{SOV}}(t+1) - q_{m}^{\text{SOV}}(t) \\& \leq \sqrt{2\Delta_{T_k}} \leq \sqrt{2T_k^2 \Phi - 2V  \sum_{t\in \mathcal{T}_k} \sum_{m\in \mathcal{S}_k} z_m^* (t) \frac{d\sigma (\zeta_m(t))}{d\zeta_m(t)}}.\notag
\end{aligned}
\end{equation}
Therefore, the energy consumption of $m \in \mathcal{S}_k$ is bounded by
\begin{equation}
\begin{aligned}
&\sum_{t\in\mathcal{T}_k} e^{\text{cm}}_m(t) + e^{\text{cp}}_{m,k}\\ &\leq E^{\text{cons}}_m + \sqrt{2T_k^2 \Phi - 2V  \sum_{t\in \mathcal{T}_k} \sum_{m\in \mathcal{S}_k} z_m^* (t) \frac{d\sigma (\zeta_m(t))}{d\zeta_m(t)}}.\notag
\end{aligned}
\end{equation}
Likewise, the energy consumption of $n \in \mathcal{U}_k$ is bounded by
\begin{equation}
\begin{aligned}
&\sum_{t\in\mathcal{T}_k} e^{\text{cm}}_n(t) \\ &\leq E^{\text{cons}}_n + \sqrt{2T_k^2 \Phi - 2V  \sum_{t\in \mathcal{T}_k} \sum_{m\in \mathcal{S}_k} z_m^* (t) \frac{d\sigma (\zeta_m(t))}{d\zeta_m(t)}}.\notag
\end{aligned}
\end{equation}
Theorem 3 is proved.
\section{Proof of Proposition 1}
\label{Proposition1}
The Lagrangian of $P3.1$ is given by:
\begin{equation}
\begin{aligned}
&\mathcal{L} = V \frac{d\sigma (\zeta_m(t))}{d\zeta_m(t)} \kappa \beta \log_2 \left(1+\frac{p_m(t)h_{m,r}(t)}{\beta N_0}\right) \\& - \kappa q_{m}^{\text{SOV}}(t) p_m(t) + \lambda_m p_m(t) + \nu_m (p^{\text{max}}_m - p_{m}(t)).\notag
\end{aligned}
\end{equation}
Then the KKT condition is given by:
\begin{equation}
\begin{aligned}
& \lambda_m^* p_m^*(t) = 0,\\
& \nu_m^* (p^{\text{max}}_m - p_{m}^*(t)) = 0,\\
& \lambda_m^*, \nu_m^*\geq 0,\\
& \frac{V \frac{d\sigma (\zeta_m(t))}{d\zeta_m(t)} \kappa \beta \frac{h_{m,r}(t)}{\beta N_0}}{1+\frac{p_m^*(t)h_{m,r}(t)}{\beta N_0}} -  q_{m}^{\text{SOV}}(t) \kappa + \lambda_m^* - \nu_m^* = 0. \notag
\end{aligned}
\end{equation}
If neither $\lambda_m^*$ nor $\nu_m^*$ is zero, there is no solution to these equations. Therefore, three cases are considered:\\
1) If $\lambda_m^* = 0$, $\nu_m^* = 0$, then
$$p_m^*(t) = \frac{V \frac{d\sigma (\zeta_m(t))}{d\zeta_m(t)} \beta}{q_{m}^{\text{SOV}}(t)} - \frac{\beta N_0}{h_{m,r}(t)}.$$\\
2) If $\lambda_m^* = 0$, $\nu_m^* \neq 0$, then
$p_m(t)^* = p^{\text{max}}_m$.\\
3) If $\lambda_m^* \neq 0$, $\nu_m^* = 0$, then $p_m(t)^* = 0$. We get:
\begin{equation}
p_m^*(t) = \left[\frac{V \frac{d\sigma (\zeta_m(t))}{d\zeta_m(t)} \beta}{q_{m}^{\text{SOV}}(t)} - \frac{\beta N_0}{h_{m,r}(t)}  \right]^{p^{\text{max}}_m}_0,\notag
\end{equation}
where $[a]^{p^{\text{max}}_m}_0$ denotes $\min(\max(a,0), p^{\text{max}}_m)$.


\begin{IEEEbiography}[{\includegraphics[width=1in,height=1.25in,clip,keepaspectratio]{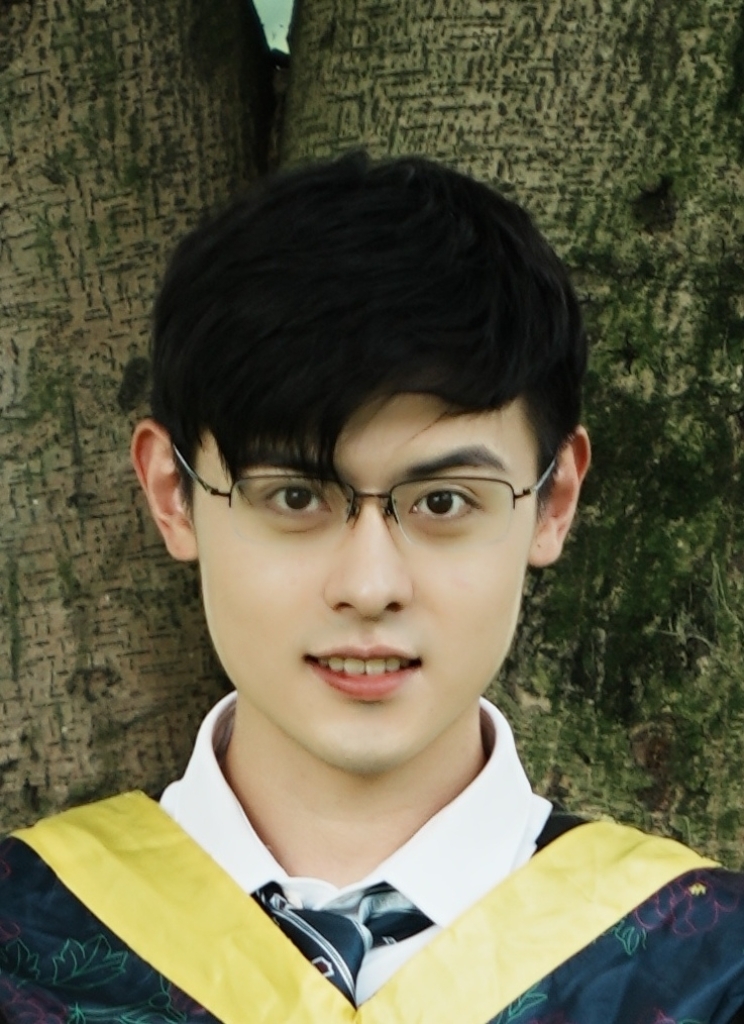}}]{Jintao Yan}(S'23)
received the B.S. degree from the University of Electronic Science and Technology of China (UESTC), Chengdu, China, in 2022. He is currently working toward the Ph.D. degree with the Network Integration for Ubiquitous Linkage and Broadband Laboratory (Niulab), Department of Electronic Engineering, Tsinghua University. His research interests include edge intelligence, vehicular networks, federated learning and optimization theory.
\end{IEEEbiography}

\begin{IEEEbiography}[{\includegraphics[width=1in,height=1.25in,clip,keepaspectratio]{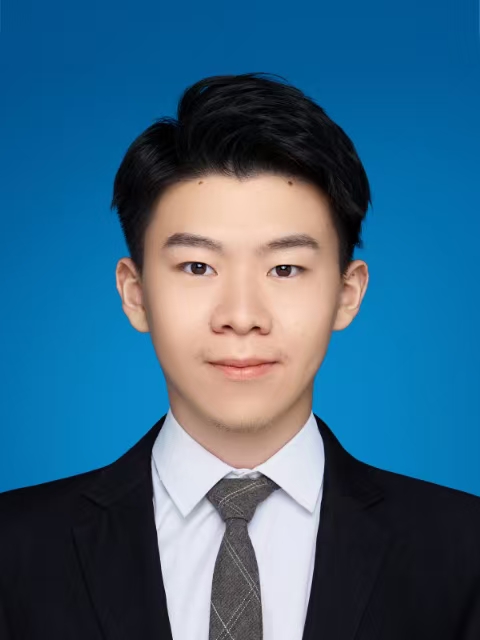}}]{Tan Chen}(S'24)
received the B.S. degree from the Department of Electronic Engineering, Tsinghua University, Beijing, China, in 2021. He is currently working toward the Ph.D. degree with the Network Integration for Ubiquitous Linkage and Broadband Laboratory (Niulab), Department of Electronic Engineering, Tsinghua University. His research interests include federated learning, vehicular networks and edge intelligence.
\end{IEEEbiography}

\begin{IEEEbiography}[{\includegraphics[width=1in,height=1.25in,clip,keepaspectratio]{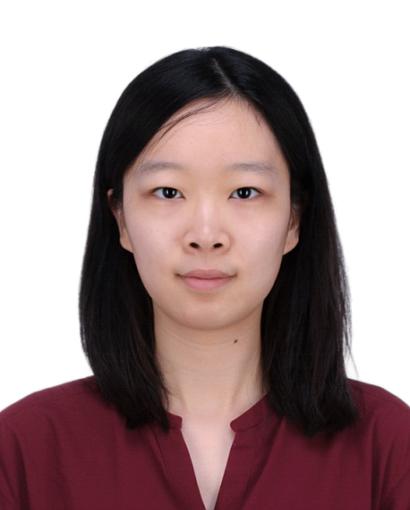}}]{Yuxuan Sun}(S'18-M'20)
received the Ph.D. degree in electronic engineering from Tsinghua University, Beijing, China, in 2020. From 2018 to 2019, she was a Visiting Student at the Department of Electrical and Electronic Engineering, Imperial College London, UK. From 2020 to 2022, she was a Post-Doctoral Researcher at the Department of Electronic Engineering, Tsinghua University, and a Visiting Researcher at Imperial College London. Currently, she is an Associate Professor at the School of Electronic and Information Engineering, Beijing Jiaotong University, Beijing, China. She is the receipt of the Young Elite Scientists Sponsorship Program by CAST.  She served as the Assistant to the Editor-in-Chief of IEEE \textsc{Transactions on Green Communications and Networking} from 2020 to 2022. She has been the Secretary of IEEE ComSoC Emerging Technologies Standing Committee since 2022. Her research interests lie in the areas of edge computing, edge intelligence, and task-oriented communications. 
\end{IEEEbiography}

\begin{IEEEbiography}[{\includegraphics[width=1in,height=1.25in,clip,keepaspectratio]{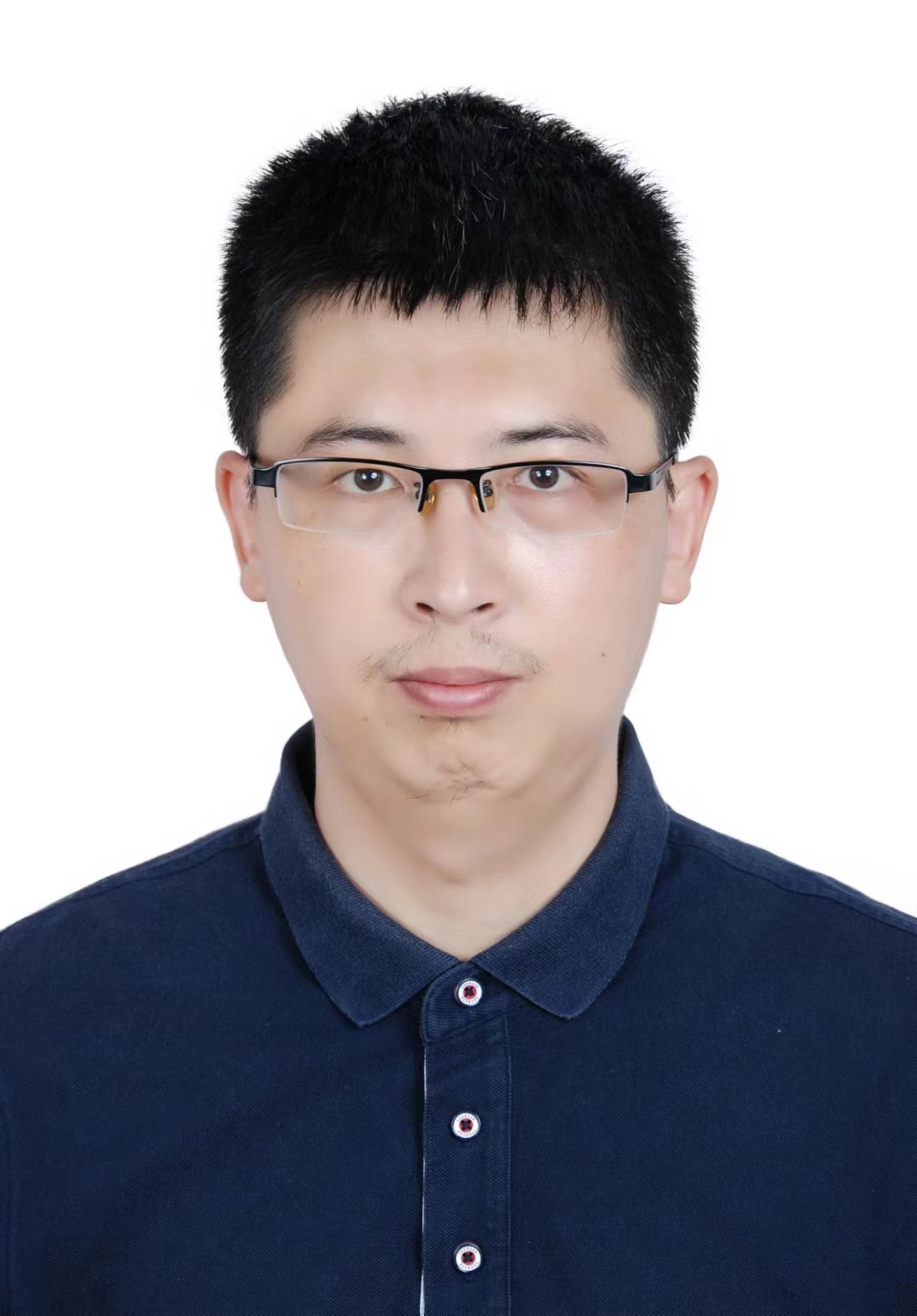}}]{Zhaojun Nan}(S’20-M’23)
received the B.S. degree in automation from Jiamusi University, Jiamusi, China, in 2009, the M.S. degree in navigation, guidance and control from Harbin Engineering University, Harbin, China, in 2012, and the Ph.D. degree in information and communication engineering from Chongqing University, Chongqing, China, in 2022. From 2012 to 2017, he worked for Tianjin 712 Communication \& Broadcasting Co., Ltd, Tianjin, China. He is currently a Postdoctoral Researcher with the Network Integration for Ubiquitous Linkage and Broadband Laboratory, Department of Electronic Engineering, Tsinghua University, Beijing, China. His research interests include mobile edge computing,  vehicular networks and autonomous driving, and green wireless communications. He has served as the TPC member for IEEE Globecom, ICC and WCNC.
\end{IEEEbiography}

\begin{IEEEbiography}[{\includegraphics[width=1in,height=1.25in,clip,keepaspectratio]{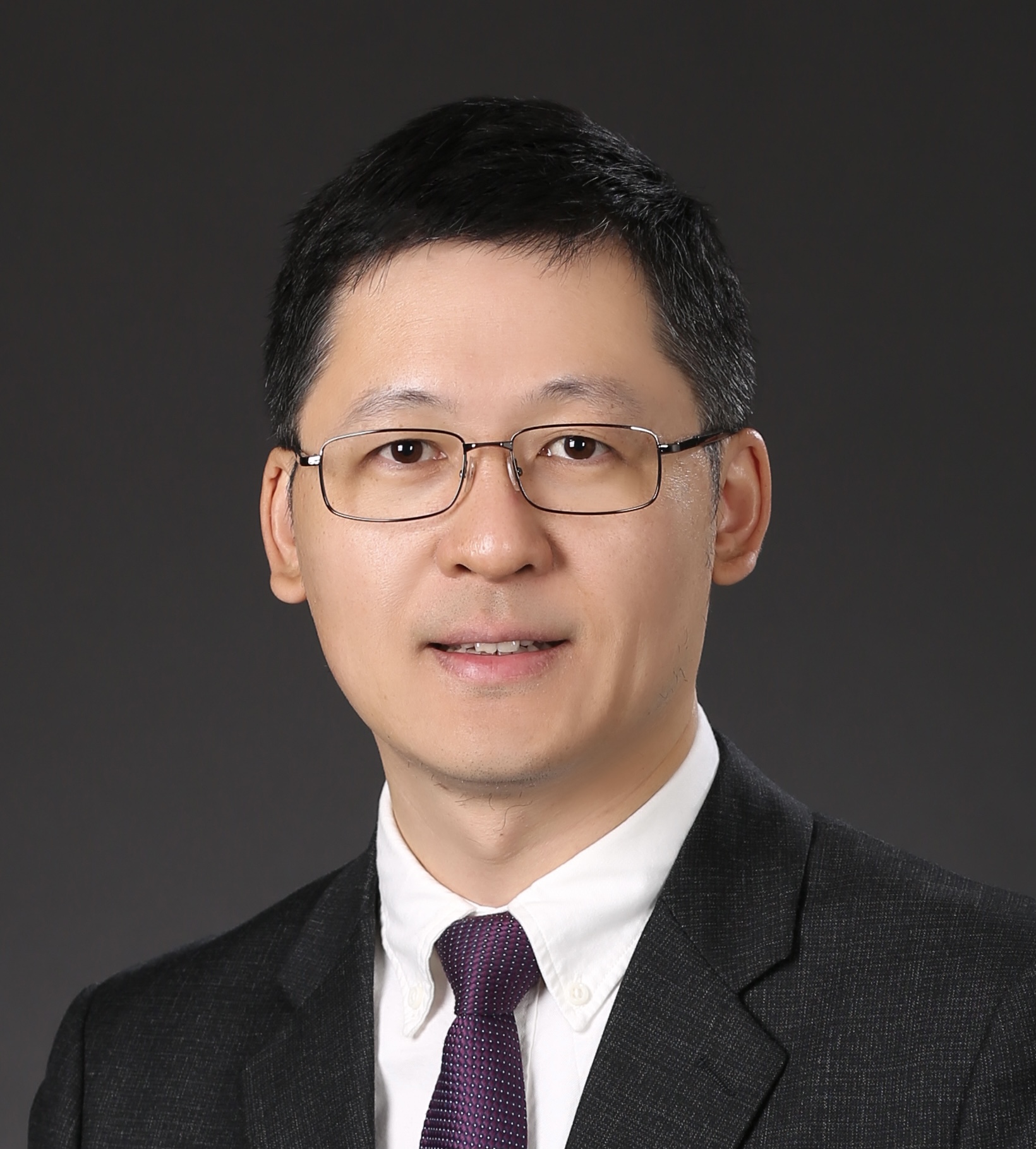}}]{Sheng Zhou} (S'06-M'12-SM'24)
received the B.E. and Ph.D. degrees in electronic engineering from Tsinghua University, Beijing, China, in 2005 and 2011, respectively. In 2010, he was a Visiting Student with the Wireless System Lab, Department of Electrical Engineering, Stanford University, Stanford, CA, USA. From 2014 to 2015, he was a Visiting Researcher with the Central Research Lab, Hitachi Ltd., Japan. He is currently an Associate Professor with the Department of Electronic Engineering, Tsinghua University. His research interests include cross-layer design for multiple antenna systems, mobile edge computing, vehicular networks, and green wireless communications. He received the IEEE ComSoc Asia–Pacific Board Outstanding Young Researcher Award in 2017, and IEEE ComSoc Wireless Communications Technical Committee Outstanding Young Researcher Award in 2020.
\end{IEEEbiography}

\begin{IEEEbiography}[{\includegraphics[width=1in,height=1.25in,clip,keepaspectratio]{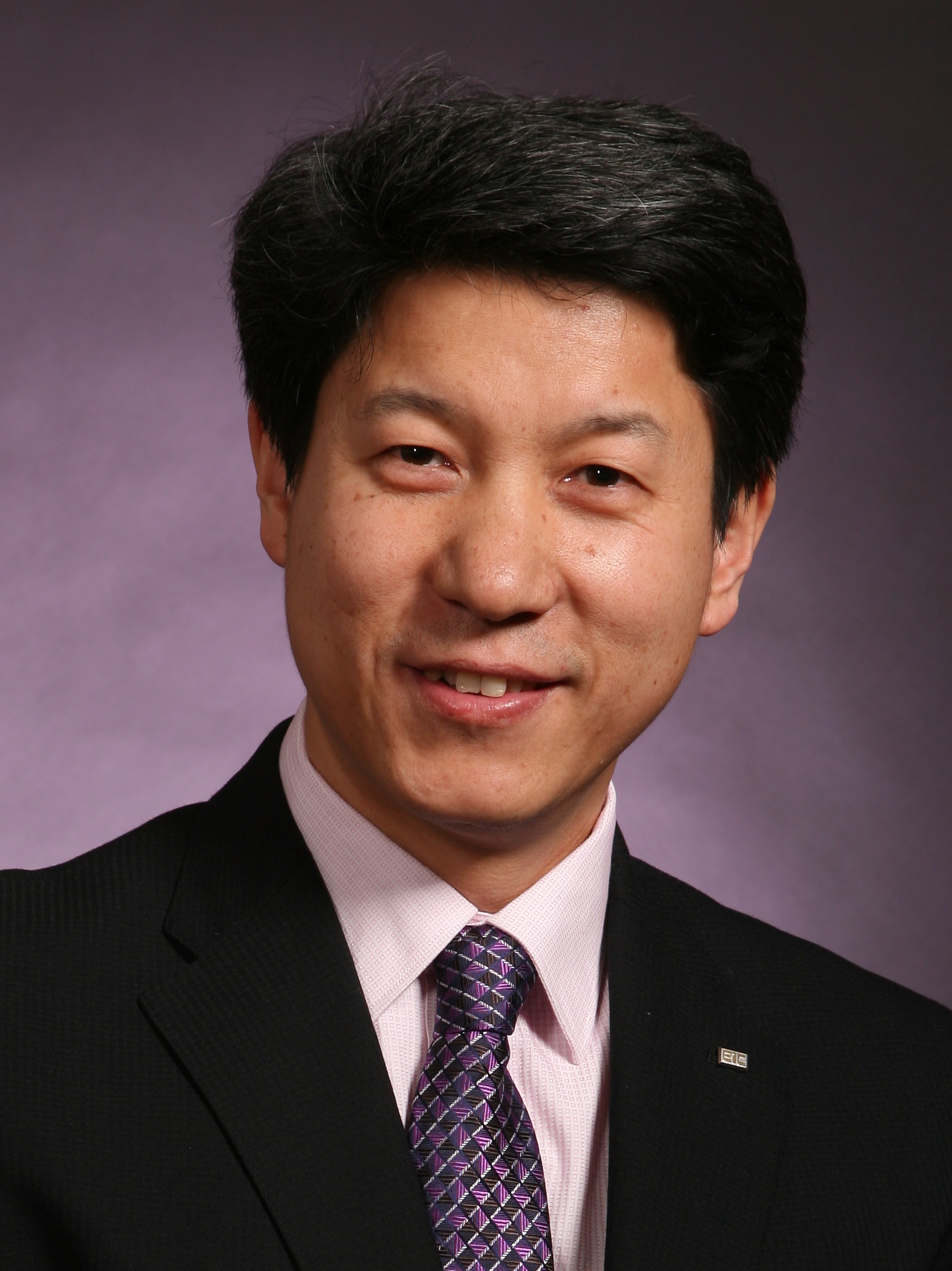}}]{Zhisheng Niu}(M'98-SM'99-F'12)
graduated from Beijing Jiaotong University, China, in 1985, and got his M.E. and D.E. degrees from Toyohashi University of Technology, Japan, in 1989 and 1992, respectively.  From 1992 to 1994, he worked for Fujitsu Laboratories Ltd., Japan, and in 1994 joined with Tsinghua University, Beijing, China, where he is now a Professor at the Department of Electronic Engineering. His major research interests include queueing theory, traffic engineering, mobile Internet, radio resource management of wireless networks, and green communication and networks.
 
Dr. Niu has been serving IEEE Communications Society since 2000 as Chair of Beijing Chapter (2000-2008), Director of Asia-Pacific Board (2008-2009), Director for Conference Publications (2010-2011), Chair of Emerging Technologies Committee (2014-2015), Director for Online Contents (2018-2019) and currently the Editor-in-Chief of IEEE \textsc{Transactions on Green Communications and Networking}. He received the Best Paper Award of Asia-Pacific Board in 2013, Distinguished Technical Achievement Recognition Award of Green Communications and Computing Technical Committee in 2018, and Harold Sobol Award for Exemplary Service to Meetings \& Conferences in 2019, all from the IEEE Communications Society. He was selected as a distinguished lecturer of IEEE Communications Society (2012-2015) as well as IEEE Vehicular Technologies Society (2014-2018). He is a fellow of both IEEE and IEICE.

\end{IEEEbiography}

\end{document}